\documentclass[runningheads]{llncs}
\usepackage{graphicx}

\usepackage{tikz}
\usepackage{comment}
\usepackage{amsmath,amssymb} 
\usepackage{color}

\usepackage{comment}
\usepackage{epsfig}
\usepackage{array,bm,multirow}
\usepackage{booktabs}
\usepackage{color}
\usepackage{url}
\usepackage{multirow}
\usepackage[lofdepth,lotdepth]{subfig}
\usepackage{caption}
\usepackage{diagbox}
\usepackage[normalem]{ulem}
\newcommand{\ccc}[1]{\textcolor{black}{#1}}

\newcommand{\os}[1]{\textcolor{black}{#1}} 

\newcommand{\omitme}[1]{}

\newcommand{\onlyarxiv}[1]{#1}
\newcommand{\app}[1]{Appendix}

\newcommand{\mvar}[2]{#1$\pm$#2} 

\usepackage[breaklinks=true,bookmarks=false,colorlinks=true]{hyperref}
\usepackage{enumitem,kantlipsum}

\begin{document}
\pagestyle{headings}
\mainmatter
\def\ECCVSubNumber{2601}  

\title{Impact of base dataset design on few-shot image classification} 

\titlerunning{Few-shot dataset design}
%
\author{
Othman Sbai\inst{1,2} \and Camille Couprie\inst{1} \and Mathieu Aubry\inst{2}
}
\authorrunning{Sbai et al.}
%

\institute{$^1$Facebook AI Research, $^2$LIGM (UMR 8049) - \'Ecole des Ponts, UPE}

\maketitle

\begin{abstract}
The quality and generality of deep image features is crucially determined by the data they have been trained on, but little is known about this often overlooked effect. In this paper, we systematically study the effect of variations in the training data by evaluating deep features trained on different image sets in a few-shot classification setting. 
The experimental protocol we define allows to explore key practical questions. What is the influence of the similarity between base and test classes? Given a fixed annotation budget, what is the optimal trade-off between the number of images per class and the number of classes? Given a fixed dataset, can features be improved by splitting or combining different classes? Should simple or diverse classes be annotated?
In a wide range of experiments, we provide clear answers to these questions on the miniImageNet, ImageNet and CUB-200 benchmarks.
We also show how the base dataset design can improve performance in few-shot classification more drastically than replacing a simple baseline by an advanced state of the art algorithm.

\end{abstract}

\section{Introduction}
\begin{figure}
    \centering
    \subfloat[{\scriptsize Annotate more classes or more examples per class?}\label{1a}]{
    \includegraphics[width=0.7\linewidth]{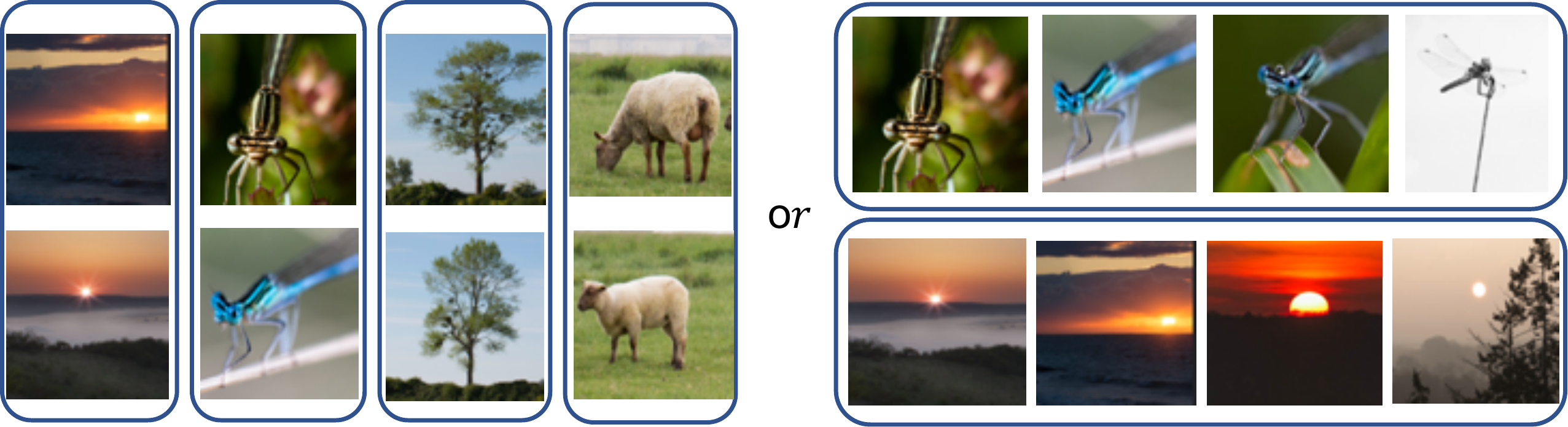}
   }\\ 
   \subfloat[{\scriptsize Build classes using more or less diverse images?}\label{1b}]{
    \includegraphics[width=0.7\linewidth]{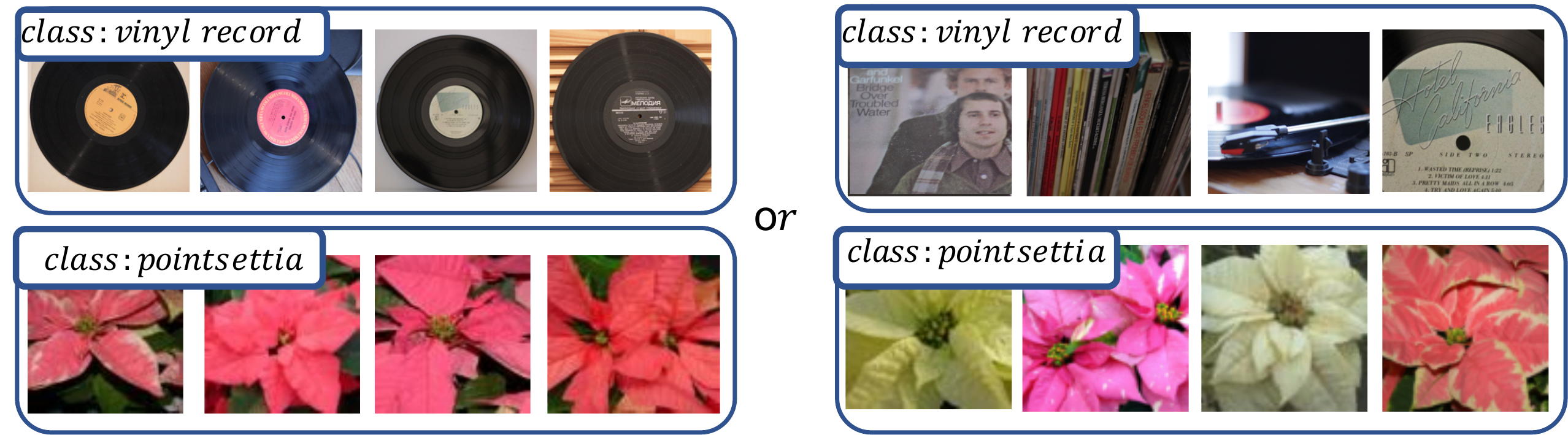}
    }
    \caption{How should we design the base training dataset and how will it influence the features?  a) Many classes with few examples / few classes with many examples; b) Simple or diverse base training images.}
    \label{fig:teaser}
\end{figure}

Deep features can be trained on a base dataset and provide good descriptors on new images \cite{sharif2014cnn,oquab2014learning}. 
The importance of large scale image annotation for the base training is now fully recognized and many efforts are dedicated to creating very large scale datasets. However, little is known on the desirable properties of such dataset, even for standard image classification tasks. To evaluate the impact of the dataset on the quality of learned features, we propose an experimental protocol based on few-shot classification. In this setting, a first model is typically trained to extract features on a base training dataset, and in a second classification stage, features are used to label images of novel classes given only few exemplars. Beyond the interest of few-shot classification itself, our protocol is well suited to vary specific parameters in the base training set and answer specific questions about its design, such as the ones presented in Fig.~\ref{fig:teaser}.

We believe this work is the first to study, with a consistent approach, the importance of the similarity of training and test data, the suitable trade-off between the number of classes and the number of images per class, the possibility of defining better labels for a given set of images, and the optimal diversity {and complexity} of the images and classes to annotate. Past studies have mostly focused on feature transfer between datasets and tasks \cite{huh2016makes,zamir2018taskonomy}. 
The study most related to ours is likely \cite{huh2016makes}, which asks the question ``What makes ImageNet good  for transfer learning?". The authors present a variety of experiments on transferring features trained on ImageNet to SUN~\cite{xiao2010sun} and Pascal VOC classification and detection~\cite{everingham2010pascal}, as well as a one-shot experiment on ImageNet.
However, using AlexNet fc7 features~\cite{krizhevsky2012imagenet}, and often relying on the WordNet hierarchy~\cite{fellbaum1998wordnet}, the authors find that variations of the base training dataset do not significantly affect transfer performance, in particular for the balance between image-per-class and classes. This is in strong contrast with our results, which outline the importance of this trade-off in our setup. We believe this might partially be due to the importance of the effect of transfer between datasets, which overshadows the differences in the learned features. 
Our few-shot learning setting precisely allows to focus on the influence of the training data without considering the complex issues of domain or task transfer.

Our work also aims at outlining data collection strategies and research directions that might lead to new performance boosts. Indeed, several works~\cite{Farabet2019lessIsMore,Triantafillou2019metadataset} have recently stressed the limitations of performance improvements brought when training on larger datasets, obtained for example by aggregating datasets~\cite{Triantafillou2019metadataset}. On the contrary, \cite{Ge_2017_CVPR} shows performance can be improved using a ``Selective Joint Fine-Tuning" strategy for transfer learning, selecting only images in the source dataset with low level feature similar to the target dataset and training jointly on both. 
Our results give insights on why it might happen, showing in particular that a limited number of images per class is often sufficient to obtain good features. Code is available at \url{imagine.enpc.fr/~sbaio/fewshot_dataset_design}.

\paragraph{Contribution.}
Our main contribution is an experimental protocol to systematically study the influence of the characteristics of the base training dataset on the resulting deep features for few-shot classification. It leads us to the following key conclusions: 
\begin{itemize}[noitemsep,topsep=0pt]
\itemsep0em 
    \item The similarity of the base training classes and the test classes has a crucial effect and standard datasets for few-shot learning consider only a very specific scenario.
    \item For a fixed annotation budget, the trade-off between the number of classes and the number of images per class has a major effect on the final performance. The best trade-off usually corresponds to much fewer images per class ($\sim 60$) than collected in most datasets.
    \item If a dataset with a sub-optimal class number is already available, we demonstrate that a performance boost can be achieved by grouping or splitting classes. While oracle features work best, we show that class grouping can be achieved using self-supervised features.
    \item {Class diversity and difficulty also have an independent influence, easier classes with lower than average diversity leading to better few-shot performances.}

\end{itemize}

\ccc{While we focus most of our analysis on a single few-shot classification approach and architecture backbone, key experiments for other methods and architectures demonstrate the generality of our results.}

\section{Related work and classical few-shot benchmarks}
\subsection{Data selection and sampling}

Training image selection is often tackled through the lens of { \bf active learning} \cite{Cohn94improvinggeneralization}. The goal of active learning is to select a subset of samples to label when training a model, while obtaining similar performance as in the case where the full dataset is annotated.  A complete review of classical active learning approaches is beyond the scope of this work and can be found in~\cite{settles2009active}. A common strategy is to remove redundancy from datasets by designing acquisition functions (entropy, mutual information, and error count)~\cite{gal2017deepactive,Farabet2019lessIsMore} to better sample training data. Specifically, \cite{Farabet2019lessIsMore} introduces an ``Adaptive Dataset Subsampling" approach designed to remove redundant samples in datasets. It predicts the
uncertainty of ensemble of models to encourage the selection of samples with
high ``disagreement". Another approach is to select samples close to the boundary decision of the model, which in the case of deep networks can be done using adversarial examples~\cite{ducoffe2018adversarial}. 
In~\cite{sener2017Active}, the authors adapt active learning strategies to batch training of neural networks and evaluate their method in a transfer
learning setting.
While these approaches select specific training samples based on their diversity or difficulty, they typically focus on performance on a fixed dataset and classes, and do not analyze performance of learned features on new classes as in our few-shot setting. 

Related to active learning is the question of online {\bf sampling strategies} to improve the training with fixed, large datasets \cite{fan2016neural,London2017adaptive,buda2017imbalance,katharopoulos2018not}. For instance, the study of \cite{buda2017imbalance} on class imbalance highlights over-sampling or under-sampling strategies that are privileged in many works. \cite{fan2016neural} and \cite{katharopoulos2018not} propose respectively reinforcement learning and importance sampling strategies to select the samples which lead to faster convergence for SGD. 

The spirit of our work is more similar to studies that try to understand key properties of good training samples to {\bf remove unnecessary samples} from large datasets. 
Focusing on the deep training process and inspired by active SVM learning approaches, \cite{vodrahalli2018all} explored using the gradient magnitude as a measure of the importance of training images. However using this measure to select training examples leads to poor performances on CIFAR and ImageNet. \cite{birodkar2019semantic} identifies redundancies in datasets such as ImageNet and CIFAR using agglomerative clustering~\cite{defays1977efficient}. Similar to us, they use features from a network pre-trained on the full dataset to compute an oracle similarity measure between the samples. However, their focus is to demonstrate that it is possible to slightly reduce the size of datasets (10\%) without harming test performance, and they do not explore further the desirable properties of a training dataset.

\subsection{Few-shot classification} 
The goal of few-shot image classification is to be able to classify images from novel classes using only a few labeled examples, relying on a large base dataset of annotated images from other classes.
Among the many deep learning approaches, the pioneer Matching networks \cite{vinyals2016matching}
and Prototypical networks \cite{snell2017prototypical} tackle the problem from a metric learning perspective. Both methods are meta-learning approaches, i.e. they train a model to learn from sampled classification episodes similar to those of evaluation. MatchingNet considers the cosine similarity to compute an attention over the support set, while ProtoNet employs an $\ell_2$ between the query and the class mean of support features.

Recently, \cite{Chen2019closer} revisited few-shot classification and showed that the simple, \ccc{meta-learning free, Cosine Classifier} baseline introduced in~\cite{gidaris2018dynamic} performs better or on par with more sophisticated approaches. Notably, its results on the CUB and Mini-ImageNet benchmarks were close to the state-of-the-art \cite{Antoniou2019sotacub,Lee_2019_CVPR}. Many more approaches have been proposed even more recently in this very active research area (e.g.~\cite{rusu2018meta,li2019finding}), including approaches relying on other self-supervised tasks (e.g. ~\cite{gidaris2019boosting}) and semi-supervised approaches
(e.g.~\cite{kim2019edge,liu2018learning,hu2019empirical}), but a complete review is outside the scope of this work, and exploration of novel methods orthogonal to our goal.

The choice of the base dataset remains indeed largely unexplored in previous studies, whereas we show that it has a huge impact on the performance, and different choices of base datasets might lead to different optimal approaches.
The Meta-dataset \cite{Triantafillou2019metadataset} study is related to our work from the perspective of analyzing dataset impact on few-shot performance. However, it investigates the effect of meta-training hyper-parameters, while our study focuses on how the base dataset design can improve few-shot classification performance. More recently,~\cite{zhou2020learning} investigates the same question of selecting base classes for few-shot learning, leading to a performance better than that of random choice, while highlighting the importance of base dataset selection in few-shot learning.

\ccc{
Since a Cosine Classifier (CC) with a Wide ResNet backbone is widely recognized as a strong baseline~\cite{gidaris2018dynamic,gidaris2019boosting,Chen2019closer,wang2019simpleshot}, we use it as reference, but also report results with two other classical algorithms, namely MatchingNet and ProtoNet.} 

\ccc{The classical benchmarks for few-shot evaluation on which we build and evaluate are listed below. Note this is not an exhaustive review, but a selection of diverse datasets which are suited to our goals.}\\

\par{{\bf Mini-ImageNet benchmark.} Mini-ImageNet is a common benchmark for few-shot learning of small resolution images~\cite{vinyals2016matching,Sachin2017}. It includes 600K images from 100 random classes sampled from the ImageNet-1K~\cite{deng2009imagenet} dataset and downsampled to $84\times84$ resolution. It has a standard split of base training, validation and test classes of 64, 16, and 20 classes respectively. 
}

\par{{\bf ImageNet benchmark.} For high-resolution images, we consider the few-shot learning benchmark proposed by \cite{hariharan2017low,wang2018low}. This benchmark splits the ImageNet-1K dataset into 389 base training, 300 validation and 311 novel classes. The base training set contains 497350 images. 
}

\par{{\bf CUB benchmark.} For fine-grained classification, we experiment with the  CUB-200-2011 dataset~\cite{wah2011caltech}. It contains 11,788 images from 200 classes, each class containing between 40 to 60 images. Following~\cite{hilliard2018few,Chen2019closer} we resize the images to $84\times84$ pixels and use the standard splits in 100 base, 50 validation and 50 novel classes and use exactly the same evaluation protocol as for mini-ImageNet.}

\section{Base dataset design and evaluation for few-shot classification}
In this section, we present the different components of our analysis. First, we explain in detail the main few-shot learning approach that we use to evaluate the influence of training data. Second, we present the large base dataset we use to sample training sets. Third, we discuss the different descriptors of images and classes that we consider, the different splitting and grouping strategies we use for dataset relabeling and the class selection methods we analyze. Finally we give details on architecture and training.

\subsection {Dataset evaluation using few-shot classification}
\label{sec:fewshot}

\begin{figure}[t]
\centering
\includegraphics[width=0.6\linewidth]{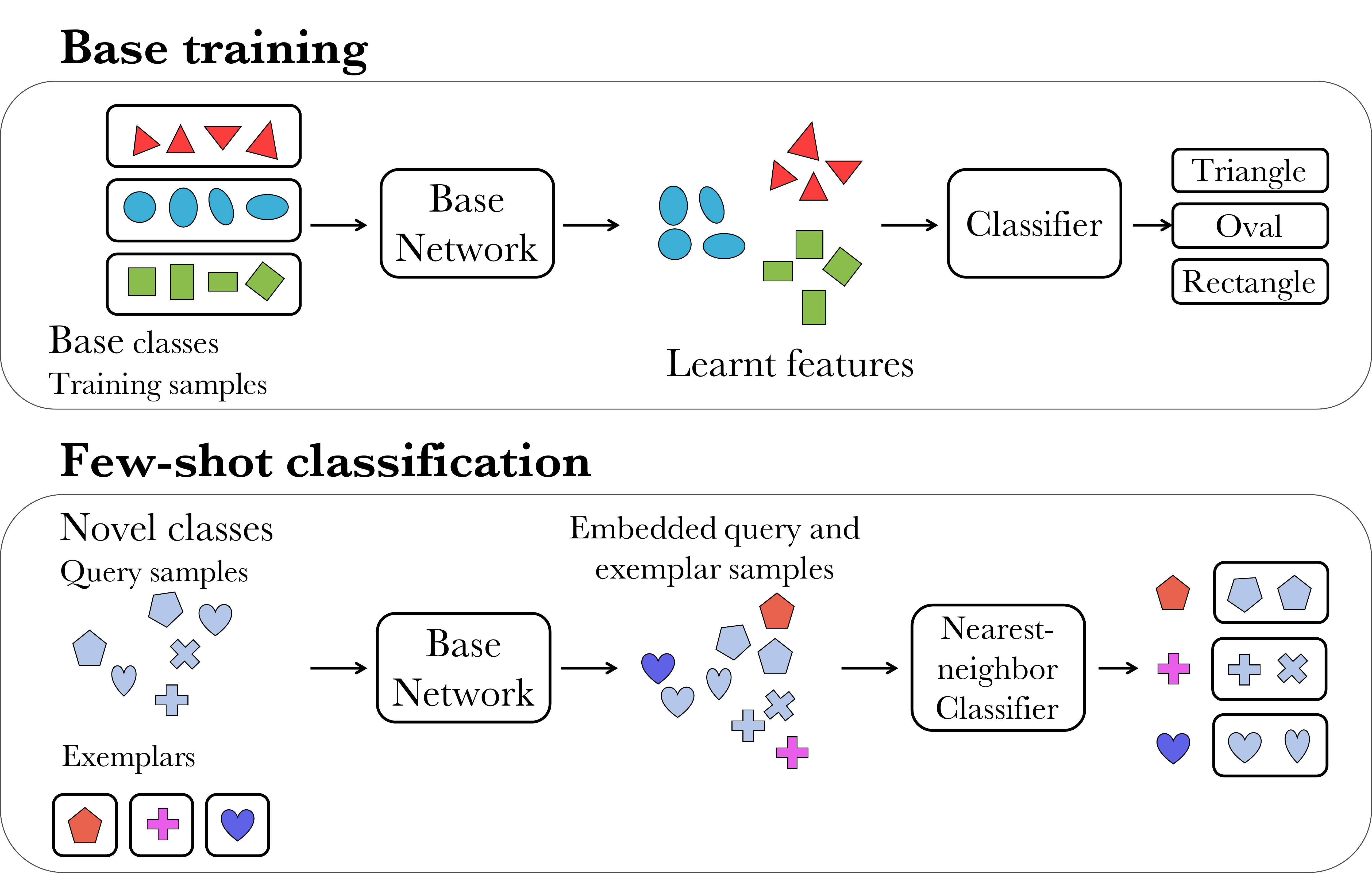}\\
\caption{Illustration of our few-shot learning framework. We train a feature extractor together with a classifier on base training classes. Then, we evaluate the few-shot classification performance of this learned feature extractor to classify novel unseen classes with few annotated examples using a nearest neighbor classifier.}

\label{fig:FSL_framework}

\end{figure}

Few-shot image classification aims at classifying test examples in novel categories using only a few annotated examples per category and typically relying on a larger base training set with annotated data for training categories. We use the simple but efficient nearest neighbor based approach, visualized in Fig.~\ref{fig:FSL_framework}.

More precisely, we start by training a feature extractor $f$ with a cosine classifier on base categories (Fig.~\ref{fig:FSL_framework} top). Then, we define a linear classifier for the novel classes as follows: if $z_i$ for $i=1...N$ are the labelled examples for a given novel class, we define the classifier weights $w$ for this class as:
\begin{equation}
\label{eq:avgFeat}
    w=\dfrac{1}{N}\sum_{i=1}^N \dfrac{f(z_i)}{\| f(z_i)\|}.
\end{equation}
In other words, we associate each test image to the novel class for which its average cosine similarity with the examples from this novel class is the highest.
Previous work on few-shot learning focuses on algorithm design for improving the classifier defined on  new labels. Instead, we explore the orthogonal dimension of base training dataset and compare the same baseline classifier using features trained on different base datasets.

\subsection{A large base dataset, ImageNet-6K}

\label{sec:datasests}

To investigate a wide variety of base training datasets, we design the ImageNet-6K dataset from which we sample images and classes for our experiments. We require both a large number of classes and a large number of images per class, to allow very diverse image selections, class splittings or groupings. We define ImageNet-6K as the subset from the ImageNet-22K dataset~\cite{ILSVRC15,deng2009imagenet} containing the largest 6K classes, excluding ImageNet-1K classes. Image duplicates are removed automatically as done in~\cite{sablayrolles2018d}. Each class has more than 900 images. For experiments on mini-ImageNet and CUB, we downsample the images to $84\times84$, and dub the resulting dataset MiniIN6K. For CUB experiments, to avoid training on classes corresponding to the CUB test set, we additionally look for the most similar images to each of the 2953 images of CUB test set using our oracle features (see Section~\ref{sec:classes}), and completely remove the 296 classes they belong to. We denote this base dataset MiniIN6K*.

\subsection{Class definition and sampling strategies}
\label{sec:classes}
\paragraph{\bf Image and class representation.}
In most experiments, we represent images by what we call {\it oracle features}, i.e. features trained on our IN6k or miniIN6K datasets. These features can be expected to provide a good notion of distance between images, but can of course not be used in a practical scenario where no large annotated dataset is available. Each class is represented by its average feature as defined in Equation~\ref{eq:avgFeat}. This class representation can be used for examples to select training classes close or far from the test classes, or to group similar classes.

We also report results with several alternative representations and metrics. In particular, we experiment with {\it self-supervised features}, which could be computed on a new type of images from a non-annotated dataset. We tried using features from  RotNet~\cite{gidaris2018unsupervised}, DeepCluster~\cite{caron2018deep}, and MoCo \cite{he2019momentum} approaches, and obtained stronger results with MoCo features which we report in the paper. \ccc{MoCo exploits the self-supervised feature clustering idea and builds a feature dictionary using a contrastive loss.} As an additionnal baseline we report results using deep features with randomly initialized weights and updated batch normalization layers during 1 epoch of miniIN6k. 
Finally, similar to several prior works, we experiment using the WordNet~\cite{fellbaum1998wordnet} hierarchy to compute similarity between classes based on the shortest path that relates their synsets and on their respective depths.

\paragraph{\bf Defining new classes.}
\label{sec:splittinggrouping}
A natural question is whether for a fixed set of images, different labels could be used to train a better feature extractor. 

Given a set of images, we propose to use existing class labels to define new classes by splitting or merging them. 
Using K-means to cluster images or classes would lead to unbalanced classes, we thus used different strategies for splitting and grouping, which we compare to K-means in the \app:
\begin{itemize}[noitemsep,topsep=0pt]
\item {\it Class splitting.} 
We iteratively split in half every class along the principal component computed over the features of the class images. We refer to this strategy as BPC (Bisection along Principal Component). 
\item {\it Class grouping.} 
To merge classes, we use a simple greedy algorithm which defines meta-classes by merging the two closest classes using their mean features, and repeat the same process for unprocessed classes recursively.
\end{itemize}

\begin{figure}[htb]
    \centering
    \setlength{\tabcolsep}{1pt}
    \fbox{
        \begin{tabular}{cccccccc}
        \includegraphics[width=0.09\linewidth]{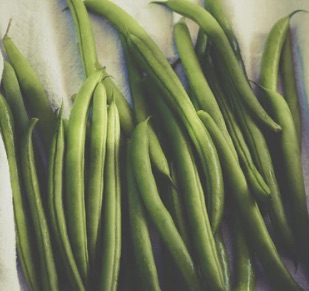}& 
        \includegraphics[width=0.09\linewidth]{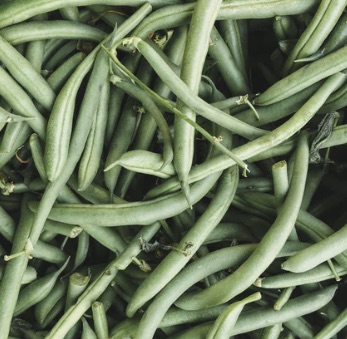}&
        \includegraphics[width=0.09\linewidth]{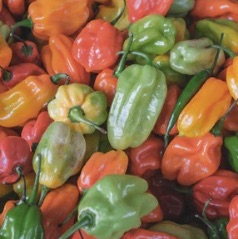}& 
        \includegraphics[width=0.09\linewidth]{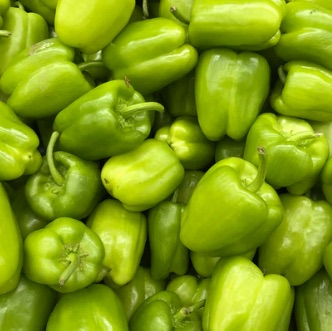}&
        \includegraphics[width=0.09\linewidth]{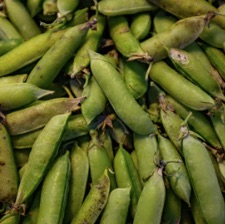}& 
        \includegraphics[width=0.09\linewidth]{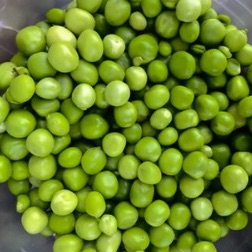}&
        \includegraphics[width=0.09\linewidth]{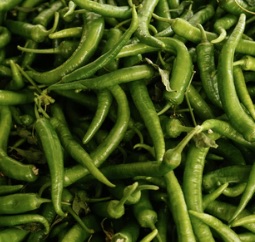}& 
        \includegraphics[width=0.09\linewidth]{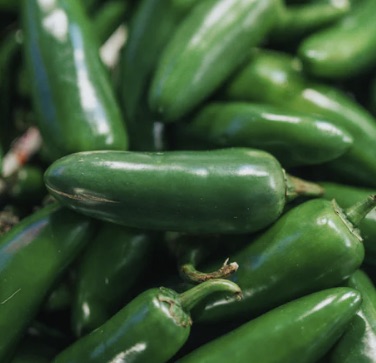}\\
        \end{tabular}
    }\\
    \fbox{
        \begin{tabular}{cccccccc}
        \includegraphics[width=0.09\linewidth]{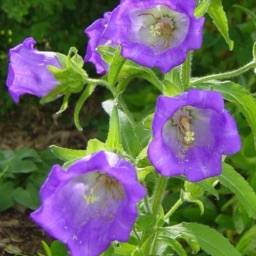}& 
        \includegraphics[width=0.09\linewidth]{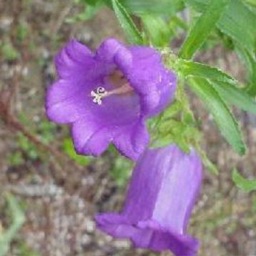}&
        \includegraphics[width=0.09\linewidth]{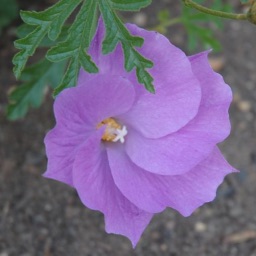}& 
        \includegraphics[width=0.09\linewidth]{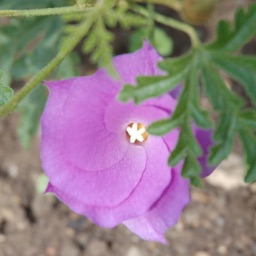}&
        \includegraphics[width=0.09\linewidth]{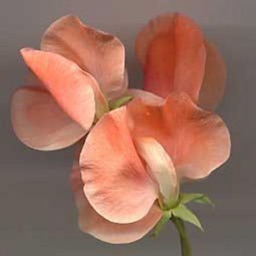}& 
        \includegraphics[width=0.09\linewidth]{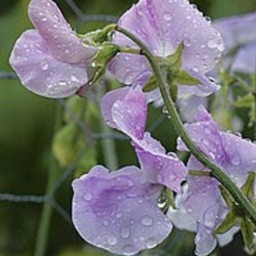}&
        \includegraphics[width=0.09\linewidth]{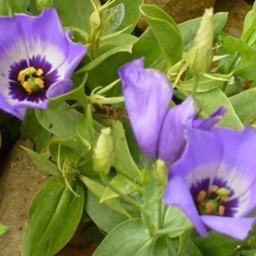}& 
        \includegraphics[width=0.09\linewidth]{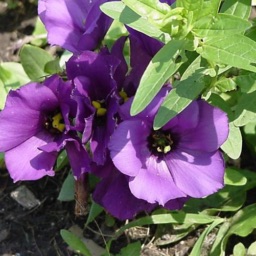}\\
        \end{tabular}
    } \\
    (a) Images from meta-classes obtained by grouping four dataset classes.\\

    \begin{tabular}{cc}
     \begin{tabular}{cccc}
        \fbox{\begin{tabular}{c}
            \includegraphics[width=0.0675\linewidth]{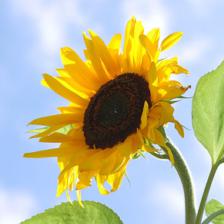} \\ \includegraphics[width=0.0675\linewidth]{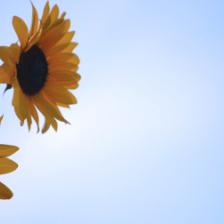}
        \end{tabular}}&
        \fbox{\begin{tabular}{c}
            \includegraphics[width=0.0675\linewidth]{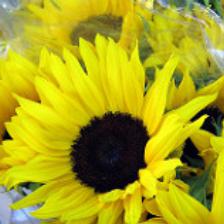}\\
            \includegraphics[width=0.0675\linewidth]{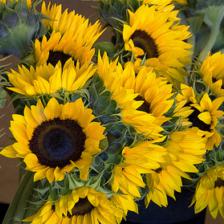}
        \end{tabular}}&
        \fbox{\begin{tabular}{c}
            \includegraphics[width=0.0675\linewidth]{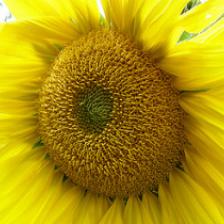}\\ 
            \includegraphics[width=0.0675\linewidth]{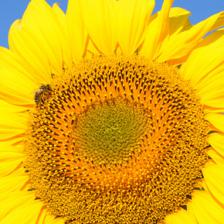}
        \end{tabular}}&
        \fbox{\begin{tabular}{c}
            \includegraphics[width=0.0675\linewidth]{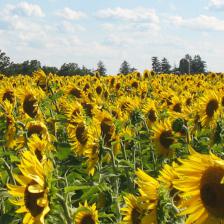}\\
            \includegraphics[width=0.0675\linewidth]{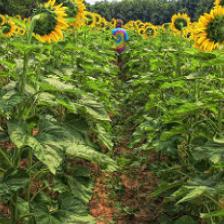}
        \end{tabular}}
    \end{tabular}
    ~~&~~
    \begin{tabular}{cccc}
        \fbox{\begin{tabular}{c}
            \includegraphics[width=0.0675\linewidth]{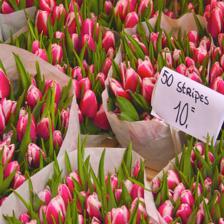}\\
            \includegraphics[width=0.0675\linewidth]{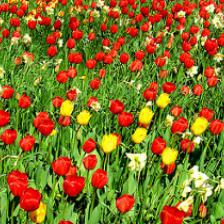}
        \end{tabular}}&
        \fbox{\begin{tabular}{c}
          \includegraphics[width=0.0675\linewidth]{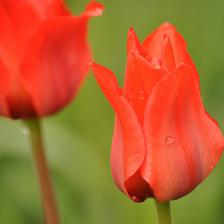}\\
          \includegraphics[width=0.0675\linewidth]{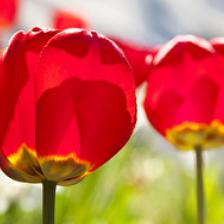}
        \end{tabular}}&
        \fbox{\begin{tabular}{c}
            \includegraphics[width=0.0675\linewidth]{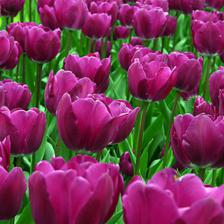}\\
            \includegraphics[width=0.0675\linewidth]{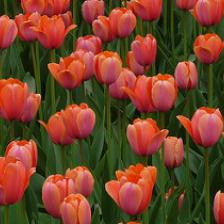}
        \end{tabular}}&
        \fbox{\begin{tabular}{c}
            \includegraphics[width=0.0675\linewidth]{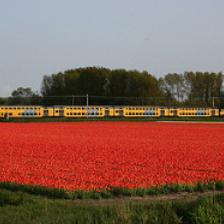}\\
            \includegraphics[width=0.0675\linewidth]{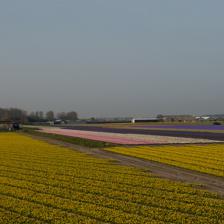}
        \end{tabular}}
        \end{tabular}
    \end{tabular}\\
    (b) Images from sub-classes obtained by splitting classes ``Sunflower" and ``Tulips".\\
    
    \begin{tabular}{cc}
     \begin{tabular}{c}
     \fbox{\begin{tabular}{cccc}
        \includegraphics[width=0.083\linewidth]{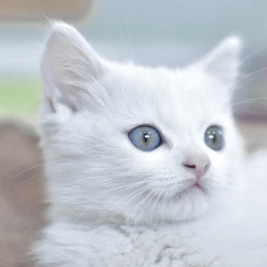}& 
        \includegraphics[width=0.083\linewidth]{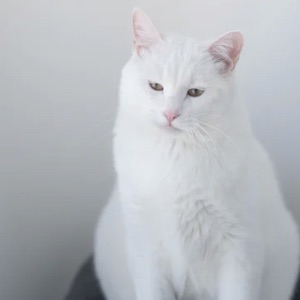}&
        \includegraphics[width=0.083\linewidth]{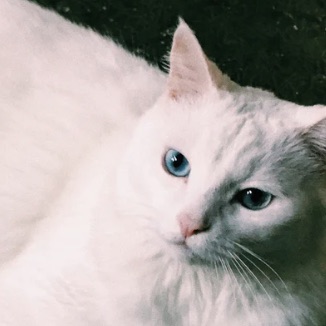}& 
        \includegraphics[width=0.083\linewidth]{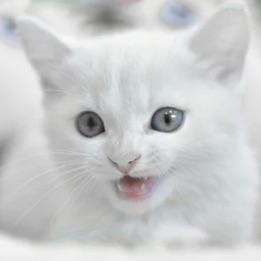}
        \end{tabular}}\\
      \fbox{\begin{tabular}{cccc}
        \includegraphics[width=0.083\linewidth]{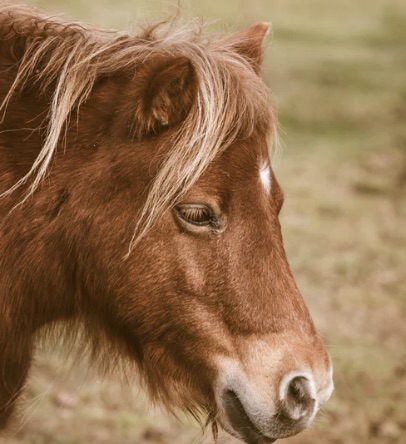}&
        \includegraphics[width=0.083\linewidth]{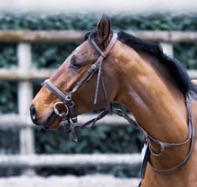}& 
        \includegraphics[width=0.083\linewidth]{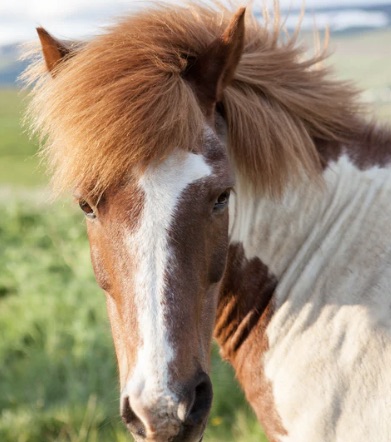}& 
        \includegraphics[width=0.083\linewidth]{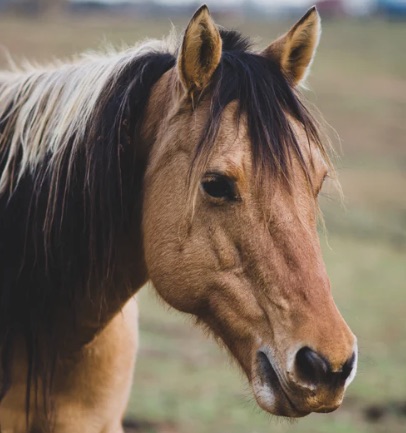}
        \end{tabular}}\\
       
        \end{tabular}&
         \begin{tabular}{c}
        \fbox{\begin{tabular}{cccc}
        \includegraphics[width=0.083\linewidth]{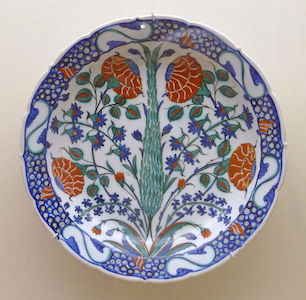}&
        \includegraphics[width=0.083\linewidth]{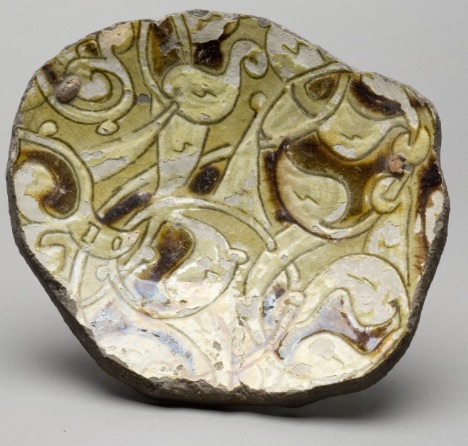}& 
        \includegraphics[width=0.083\linewidth]{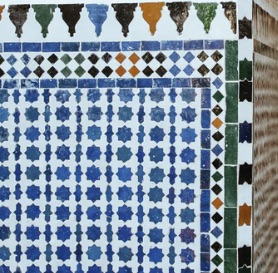}& 
        \includegraphics[width=0.083\linewidth]{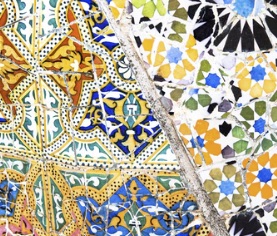}

        \end{tabular}}\\
        \fbox{\begin{tabular}{cccc}
        \includegraphics[width=0.083\linewidth]{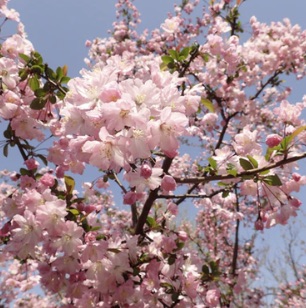}& 
        \includegraphics[width=0.083\linewidth]{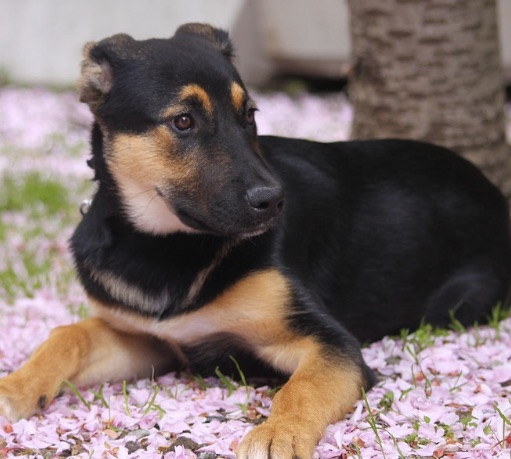}&
        \includegraphics[width=0.083\linewidth]{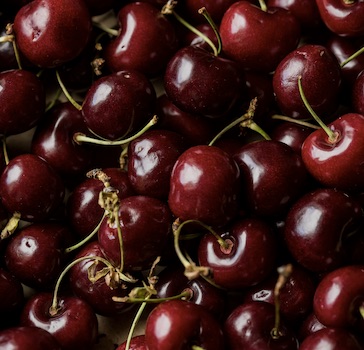}& 
        \includegraphics[width=0.083\linewidth]{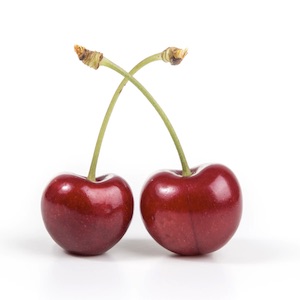}\\
      \end{tabular}}
        
        \end{tabular}\\
        c) Least diverse classes  & d) Most diverse classes \\
        (Horse, Angora cat ...) &  (Arabesque, cherry...) 
    \end{tabular}

    \caption{a) Images from {\bf meta-classes obtained by grouping} dataset classes using pre-trained features. Each line represents a meta-class. b) Examples of \textbf{sub-classes obtained by splitting} dataset classes using pre-trained features. Each column represents a sub-class. c), d) Images from least or most diverse classes from miniIN6k, with one line per class.}
    \label{fig:class_splitting}
\end{figure}

We display examples of resulting grouped and split classes in Figure~\ref{fig:class_splitting}.

\paragraph{\bf Measuring class diversity and difficulty.}

\label{sec:div}
One of the questions we ask is whether class diversity impacts the trained features' few-shot performance. We therefore analyze results by sampling classes more or less frequently according to their diversity and difficulty:
\begin{itemize}[noitemsep,topsep=0pt]
    \item {\it Class diversity.} We use the variance of the normalized features as a measure of class diversity.\onlyarxiv{ We show in Figure \ref{fig:class_splitting}~(c,d) examples of least and most diverse classes.} Classes with low feature variance consist of very similar looking objects or simple visual concepts while the ones with high feature variance represent abstract concepts or include very diverse images. 
    \item {\it Class difficulty.} To measure the difficulty of a class, we use the validation accuracy of our oracle classifier. 
\end{itemize}

\subsection{Architecture and training details}

We use different architectures and training methods in our experiments. Similar to previous works~\cite{wang2019simpleshot,Chen2019closer}, we employ WRN28-10, ResNet10, ResNet18 and Conv4 architectures. The ResNet architectures are adapted to handle $84\times 84$ images by replacing the first convolution with a kernel size of 3 and stride of 1 and removing the first max pooling layer. In addition to the cosine classifier described in Section
~\ref{sec:fewshot}, we experiment with the classical Prototypical Networks~\cite{snell2017prototypical} and Matching Networks~\cite{vinyals2016matching}. 

Since we compare different training datasets, we adapt the training schedule depending on the size of the training dataset and the method. For example on MiniIN-6k, we train Prototypical Networks and Matching Networks for 150k episodes, while when training on smaller size datasets we use 40k episodes as in~\cite{Chen2019closer}. We use fewer query images per class when training on classes with not enough images per class for Prototypical and Matching Networks.

When training a Cosine Classifier, we train using an SGD optimizer with momentum of 0.9 and weight decay of $5.10^{-4}$ for 90 epochs starting with an initial learning rate of 0.05 and dividing it by 10 every 30 epochs. We also use a learning rate warmup for the first 6K iterations, that we found beneficial for stabilizing the training and limiting the variance of the results. For large datasets with more than $10^6$ images, we use a batch size of 256 and 8 GPUs to speed up the training convergence, while for smaller datasets (most of our experiments are done using datasets of 38400 images, as in MiniIN training set), we use a batch size of 64 images and train on a single GPU. During training, we use a balanced class sampler that ensures sampled images come from a uniform distribution over the classes regardless of their number of images. 

On the ImageNet benchmark, we use a ResNet-34 network and trained for 150K dividing the learning rate by 10 after 120K, 135K and 145K iterations using a batch size of 256 on 1 GPU.

Following common practices, during evaluation, we compute the average top-1 accuracy on 15 query examples over 10k episodes sampled from the test set on 5-way tasks for miniIN and CUB benchmarks, while we compute the top-5 accuracy on 6 query examples over 250-way tasks on the ImageNet benchmark.

\section{Analysis}
\subsection{Importance of base data and its similarity to test data}
\label{sec:sim}

\begin{table}[tbp]
\begin{center}

\resizebox{\textwidth}{!}{\begin{tabular}[htp]{cc|c|c|c|c|c|c|}
\cline {3-8}
& &  \multicolumn{3}{c|}{ MiniIN test} &\multicolumn{3}{c|}{CUB test }\\
\cline {2-8} 
& \multicolumn{1}{|c|}{\backslashbox{Algo.}{Base\\data}}& \begin{tabular}{@{}c@{}} MiniIN \\ $N$=38400 \\ $C=64$ \end{tabular} & \begin{tabular}{@{}c@{}} MiniIN6K\\ Random\\$N$=38400  \\$C=64$ \end{tabular} & \begin{tabular}{@{}c@{}} MiniIN6K \\$N$$\approx$7,1.$10^6$\\ $C$=6000 \end{tabular} & \begin{tabular}{@{}c@{}} CUB \\  $N$=5885  \\ $C=100$ \end{tabular} & \begin{tabular}{@{}c@{}} MiniIN6K*\\ Random \\ $N$=38400  \\ $C=64$ \end{tabular} & \begin{tabular}{@{}c@{}} MiniIN6K* \\$N$$\approx$6,8.$10^6$\\ $C=5704$ \end{tabular} \\
\hline 
\multicolumn{1}{|l|}{}& PN~\cite{snell2017prototypical} &
   \mvar{73.64}{0.84}  & \mvar{70.26}{1.30} &  \mvar{85.14}{0.28} & \mvar{87.84}{0.42}  & \mvar{52.51}{1.57} &  \mvar{68.62}{0.5} \\
\cline {2-8}
 \multicolumn{1}{|l|}{WRN}& MN~\cite{vinyals2016matching} &
 \mvar{69.19}{0.36}  & \mvar{65.45}{1.87} &  \mvar{82.12}{0.27} & \mvar{85.08}{0.62}  & \mvar{46.32}{0.72} & \mvar{59.90}{0.45} \\
 \cline {2-8}
 \multicolumn{1}{|l|}{}& CC &
 \mvar{\bf 78.95}{0.24}  & \mvar{\bf 75.48}{1.53} &  \mvar{\textbf{96.91}}{0.14} & \mvar{\textbf{90.32}}{0.14}  & \mvar{\bf 58.03}{1.43} &  \mvar{\textbf{90.89}}{0.10} \\ 
 \hline
 \multicolumn{1}{|l|}{Conv4}& CC &
  \mvar{65.99}{0.04}  & \mvar{64.05}{0.75} &  \mvar{74.56}{0.12} & \mvar{80.71}{0.15}  & \mvar{56.44}{0.63} &  \mvar{66.81}{0.30} \\
\hline
\multicolumn{1}{|l|}{ResNet10}& CC &
\mvar{76.99}{0.07}  & \mvar{74.17}{1.42} & \mvar{91.84}{0.06} & \mvar{89.07}{0.15} & \mvar{57.01}{1.44}  & \mvar{82.20}{0.44} \\
\hline
\multicolumn{1}{|l|}{ResNet18}& CC &
\mvar{78.29}{0.05}  & \mvar{75.14}{1.58} &  \mvar{93.36}{0.19} & \mvar{89.99}{0.07}  & \mvar{56.64}{1.28} &  \mvar{88.32}{0.23} \\
\hline
\end{tabular}}
\end{center}
\caption{5-shot, 5-way accuracy on MiniIN and CUB test sets using different base training data, algorithms and backbones. PN: Prototype Networks~\cite{snell2017prototypical}. MN: Matching Networks~\cite{vinyals2016matching}. CC: Cosine Classifier. WRN: Wide ResNet28-10. MiniIN6K (resp. MiniIN6K*) Random: 600 images from 64 classes sampled randomly from MiniIN6K (resp. MiniIN6K*). We evaluate the variances over 3 different runs.
}

\label{tab:table1}
\end{table}

We start by validating the importance of the base training dataset for the few-shot classification, both in terms of size and of the selection of classes.
In Table~\ref{tab:table1}, we report five shot results on the CUB and MiniIN datasets, the one shot results are available in the \app\onlyarxiv{ in Table~\ref{tab:table1_1shot}}. We write $N$ the total number of images in the dataset and $C$ the number of classes. {Similar results on ImageNet benchmark can be read in \onlyarxiv{Fig.~\ref{fig:tradeoff_IN} of }the }\app. On the miniIN benchmark,  we observe that our implementation of the strong CC baseline \ccc{using a WRN backbone} yields slightly better performance using miniIN base classes than the ones reported in~\cite{gidaris2019boosting,Lee_2019_CVPR}(76.59). \ccc{We validate the consistency of our observations by varying algorithms and architectures using the codebase of \cite{Chen2019closer}.} 

Our first finding is that using the whole miniIN-6K dataset for the base training boosts the performance on miniIN by a very large amount, 20\% and 18\% for 1-shot and 5-shot classification respectively, compared to training on 64 miniIN base classes. Training on IN-6K images also results in a large $10\%$ boost in 5-shot top-5 accuracy on ImageNet benchmark.
Another interesting result is that sampling random datasets of 64 classes and 600 images per class leads to a 5-shot performance of $75.48\%$ on MiniIN clearly below the one using the base classes from miniIN $78.95\%$. \ccc{A similar  observation can be made for different backbones (Conv4, ResNets) and algorithms tested (ProtoNet, MatchingNets),} as well as on the ImageNet benchmark.
A natural explanation for these differences is that the base training classes from the benchmarks are correlated to the test classes. 

\begin{figure}[t]
\begin{tabular}{cc}
\includegraphics[width=0.48\linewidth]{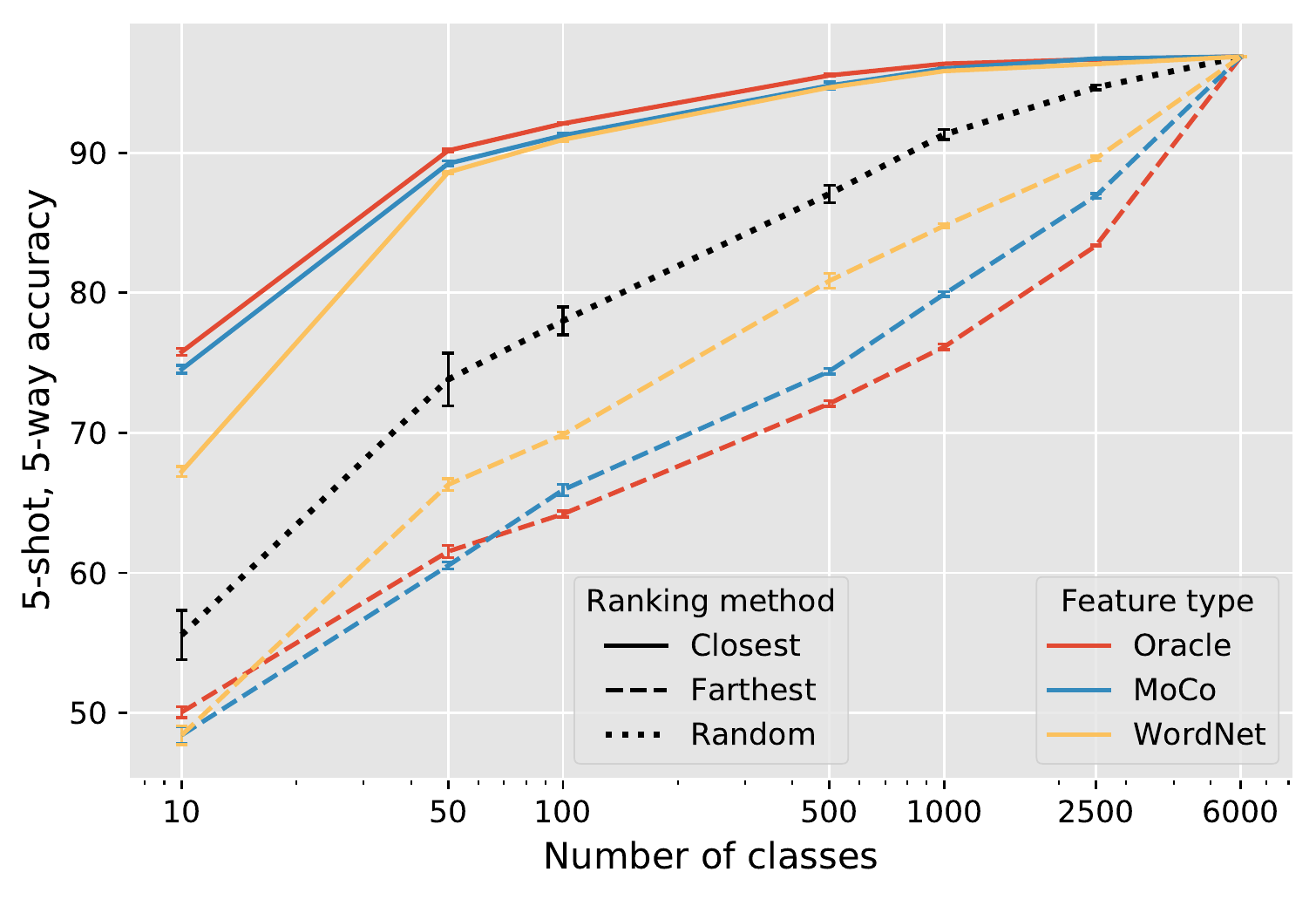} & 
\includegraphics[width=0.48\linewidth]{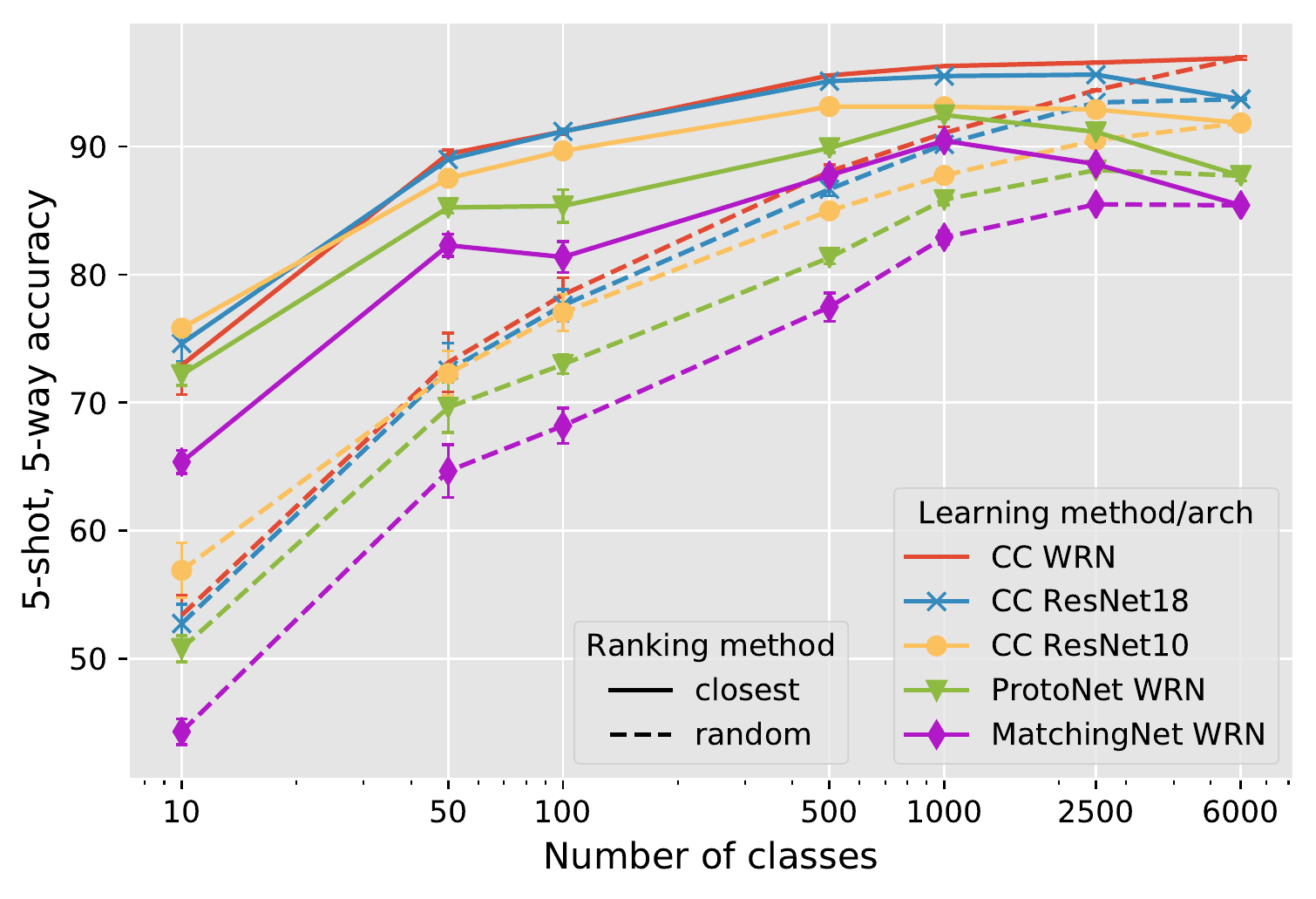} \\
(a) Different selection criteria & (b) Different backbones and algorithms \\
\end{tabular}

 \caption{Five-shot accuracy on miniIN when sampling classes from miniIN-6K {randomly or closest/farthest} to the miniIN test set using 900 images per class. (a) Comparison between different class selection criteria for selecting classes closest or farthest from the test classes. (b) Comparison of results with different algorithms and backbones using oracle features to select closest classes. }
 \label{fig:adding_classes}
\end{figure}

To validate this hypothesis, we selected a varying number of base training classes from miniIN-6K closest and farthest to miniIN test classes using either oracle features, MoCo features, or the WordNet hierarchy, and report the results of training using a cosine classifier with WRN architecture in Fig.~\ref{fig:adding_classes}a. A similar experiment on CUB is shown \onlyarxiv{in Fig.~\ref{fig:cub_similarity_features} }in the \app. We use 900 random images for each class.
While all features used for class selection yield similarly superior results for closest class selection and worst results for farthest class selection, we observe that using oracle features leads to larger differences than using MoCo features and Wordnet hierarchy. 
In Fig.~\ref{fig:adding_classes}b, we study the influence of the architecture and training method on the previously observed importance of class similarity to test classes. Similar gaps can be observed in all cases. Note however that for ProtoNet, MatchingNet and smaller backbones with CC, the best performance is not obtained with the largest number of classes.

While these findings themselves are not surprising, the amplitude of performance variations demonstrates the importance of studying the influence of training data and strategies for data selection, especially considering that most advanced few-shot learning strategies only increase performance by a few points compared to strong nearest neighbor based baselines such as CC \cite{Chen2019closer,qiao2018few}.

\subsection{Effect of the number of classes for a fixed number of annotations}

\begin{figure}[tb!]
\begin{center}
\begin{tabular}{cc}
    \includegraphics[width=0.48\linewidth]{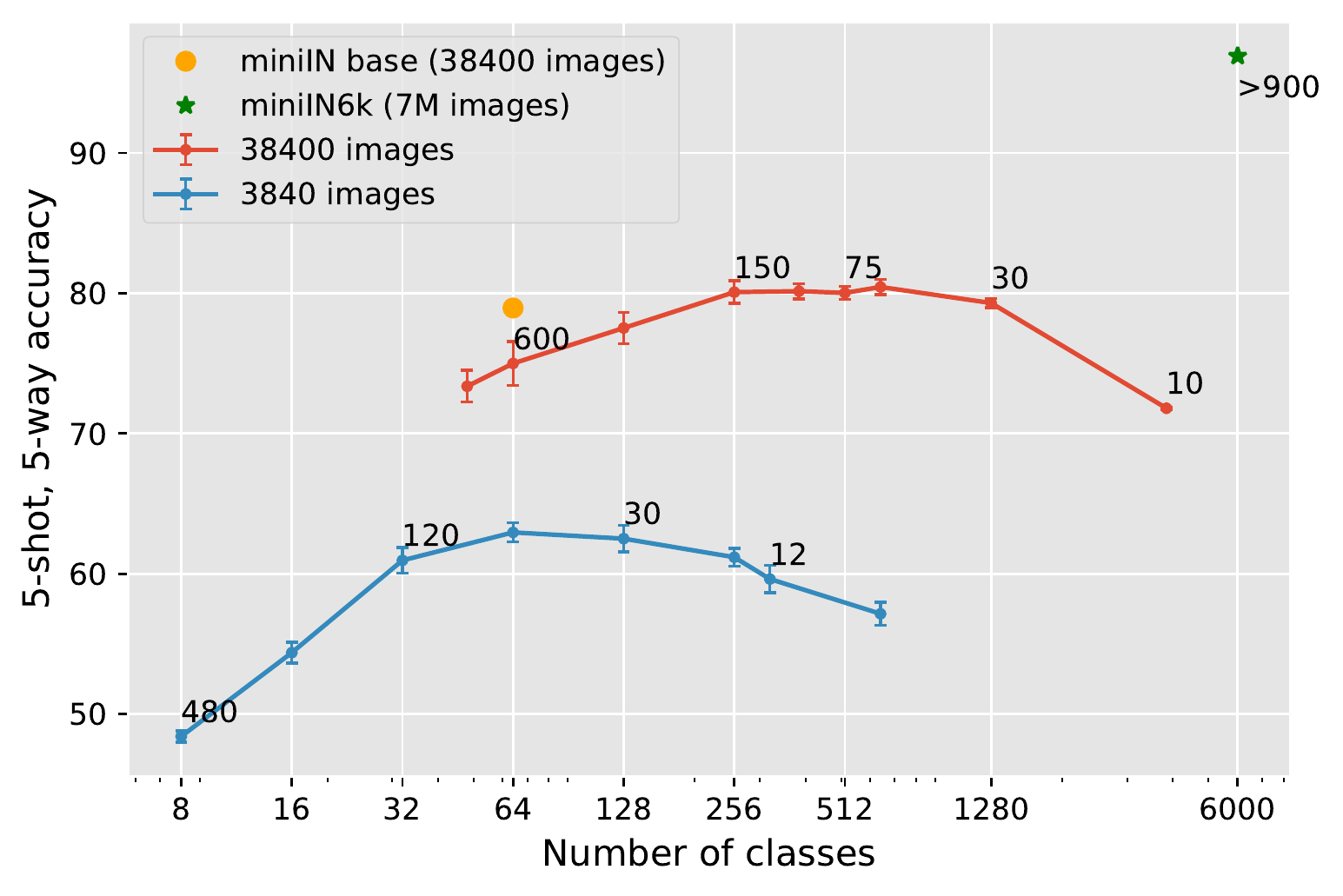} & 
     \includegraphics[width=0.48\linewidth]{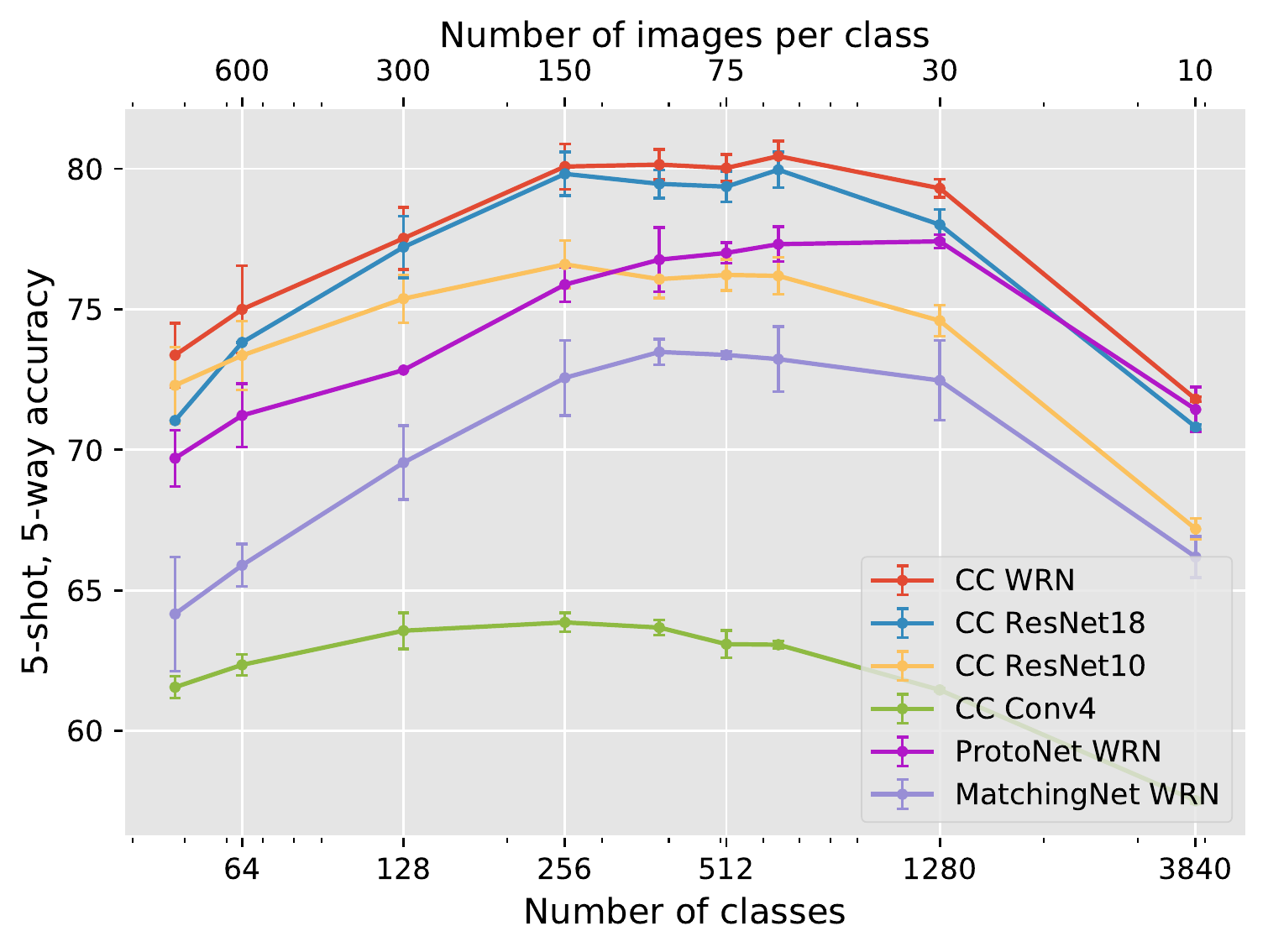}\\
    a) Different budgets on MiniIN & b) Different setups on MiniIN\\
 \onlyarxiv{ \includegraphics[width=0.48\linewidth]{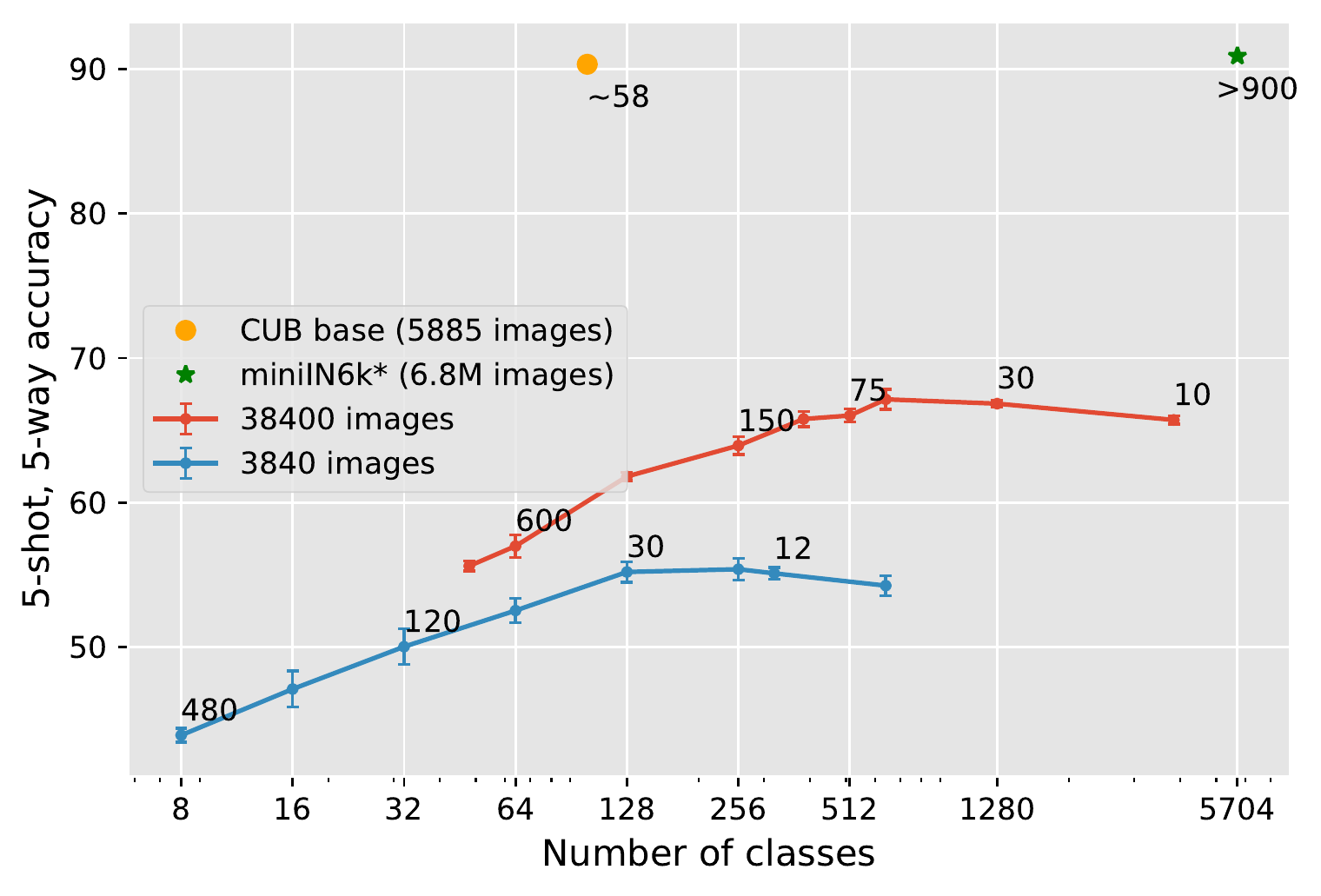}&
 \includegraphics[width=0.48\linewidth]{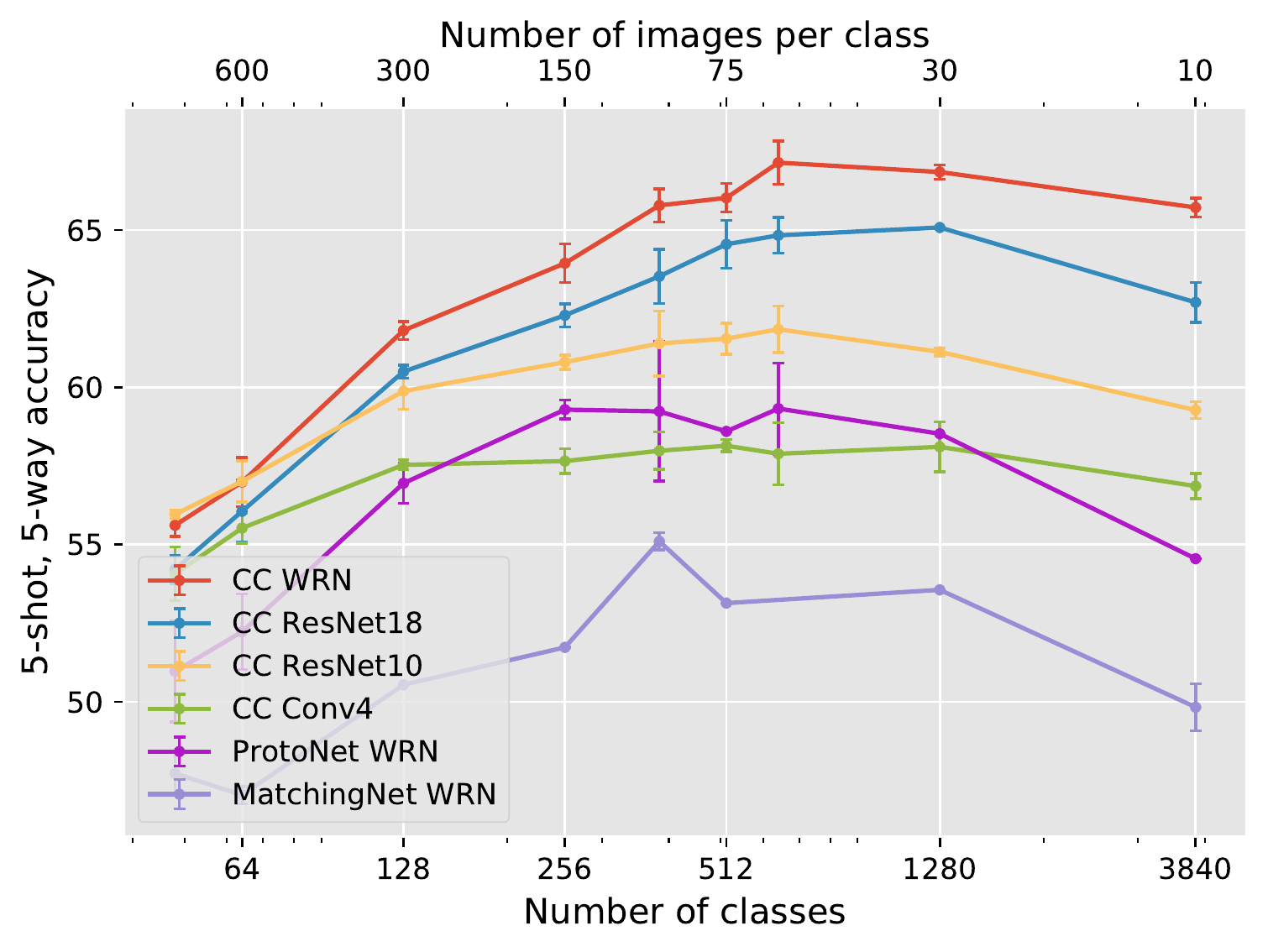} \\
 c)  Different budgets on CUB & d) Different setups on CUB \\}
 \end{tabular}
 \end{center}

 \caption{Trade-off between the number of classes and images per class for a fixed image budget. In (a\onlyarxiv{,c}) we show the trade-off for different dataset sizes and points are annotated with the corresponding number of images per class. In (b\onlyarxiv{,d}) we consider a total budget of 38400 annotated images and show the trade-off for different architectures and methods. The top scale shows the number of images per class and the bottom scale the number of classes.
 }
 \label{fig:tradeoff}
\end{figure}

An important practical question when building a base training dataset is the number of classes and images to annotate, since the constraint is often the cost of the annotation process. We thus consider a fixed number of annotated images and explore the effect of the trade-off between the number of images per class and the number of classes. In Fig.~\ref{fig:tradeoff}, we visualize the 5-shot performance resulting from this trade-off in the base training classes on the miniIN and CUB benchmarks. In all cases, we select the classes and images randomly from our miniIN6K and miniIN6k* dataset respectively, and plot the variance over 3 runs.  

First, in Fig.~\ref{fig:tradeoff} (a,b) we compare the trade-off for different numbers of annotated images. We sample randomly datasets of 38400 or 3840 images with different number of classes and the same number of image in each class. We also indicate the performance with the standard benchmarks base dataset and the full miniIN6K data. The same graph on ImageNet benchmark can be seen in\onlyarxiv{ Fig.~\ref{fig:tradeoff_IN} of} the \app~using 50k and 500k images datasets.
 
As expected, the performance decreases when too few classes or too few images per classes are available. Interestingly, on the miniIN test benchmark (Fig.~\ref{fig:tradeoff}a) the best performance is obtained around 384 classes and 100 images per class with a clear boost (around $5\%$) over the performance using 600 images for 64 classes which is the trade-off chosen in the miniIN benchmark\onlyarxiv{. In Fig.~\ref{fig:tradeoff}b}, we observe that the best trade-off is very different on the CUB benchmark, corresponding to more classes and very few images per class. We believe this is due to the fine-grained nature of the dataset.

Second, in Fig.~\ref{fig:tradeoff} (b\onlyarxiv{,d}), we study the consistency of these findings for different architectures and few-shot algorithms with a 38400 annotated images budget. While the trade-off depends on the architecture and method, there is always a strong effect, and the optimum tends to correspond to much fewer images per class than in standard benchmarks. For example, the best performance with ProtoNet and MatchingNet on the miniIN benchmark is obtained with as few as 30 images per class. This is interesting since it shows that the ranking of different few-shot approaches may depend on the trade-off between number of base images and classes selected in the benchmark. 

The importance of this balance, and the fact that it does not necessarily correspond to the one used in the standard datasets is also important if one wants to pre-train features with limited resources. Indeed, better features can be obtained by using more classes and less images per class compared to using all available images for the classes with the largest number of images as is often done, with the idea to avoid over-fitting. Again, the boost observed for few-shot classification performance is very important compared to the ones provided by many advanced few-shot learning approaches.

\begin{figure}[htb]
    \centering
    \begin{tabular}{cc}
    \includegraphics[width=0.48\linewidth ]{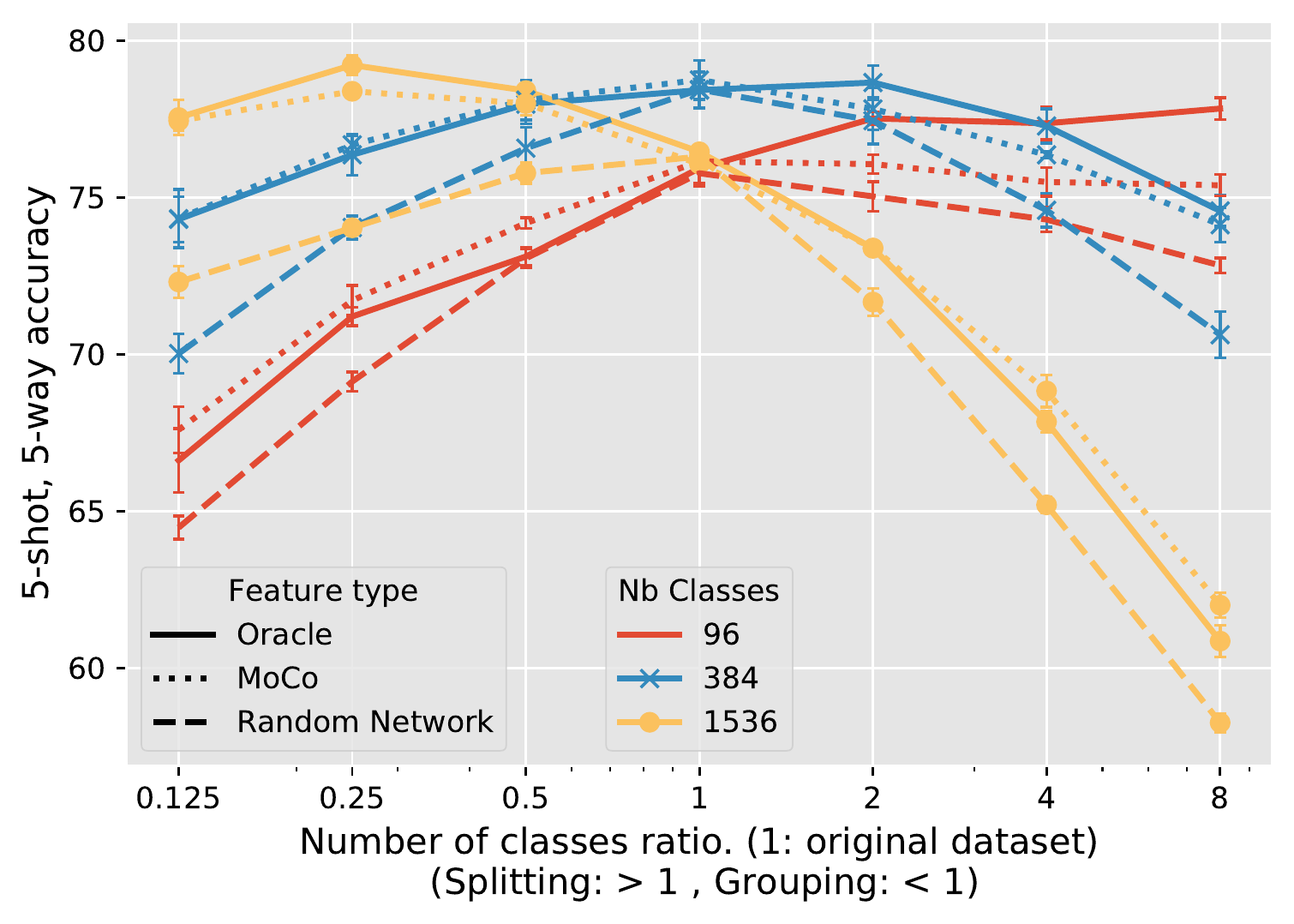}& 
    \includegraphics[width=0.48\linewidth ]{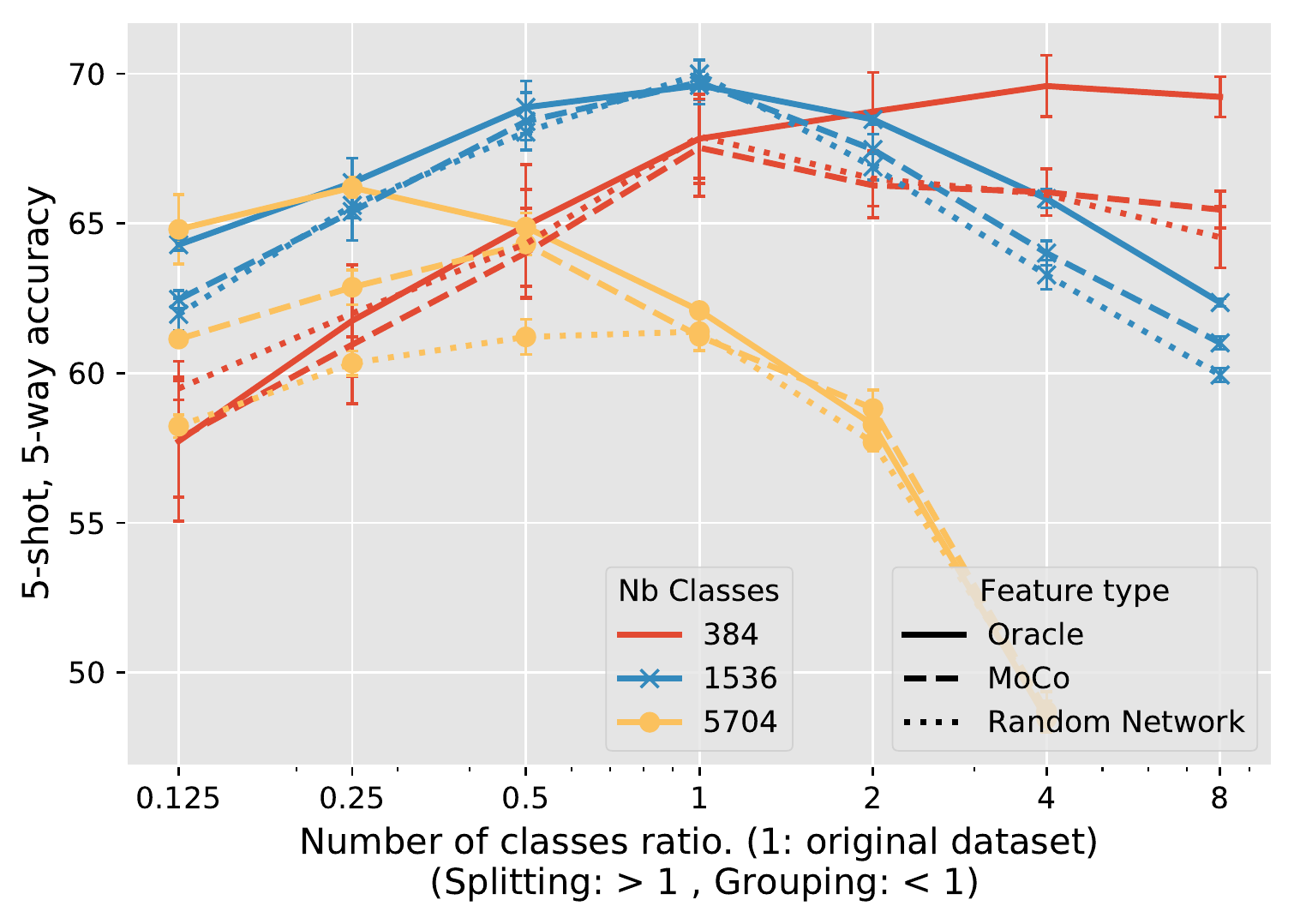}\\
     a) On MiniIN, using different features & b) On CUB, using different features\\
    \includegraphics[width=0.48\linewidth ]{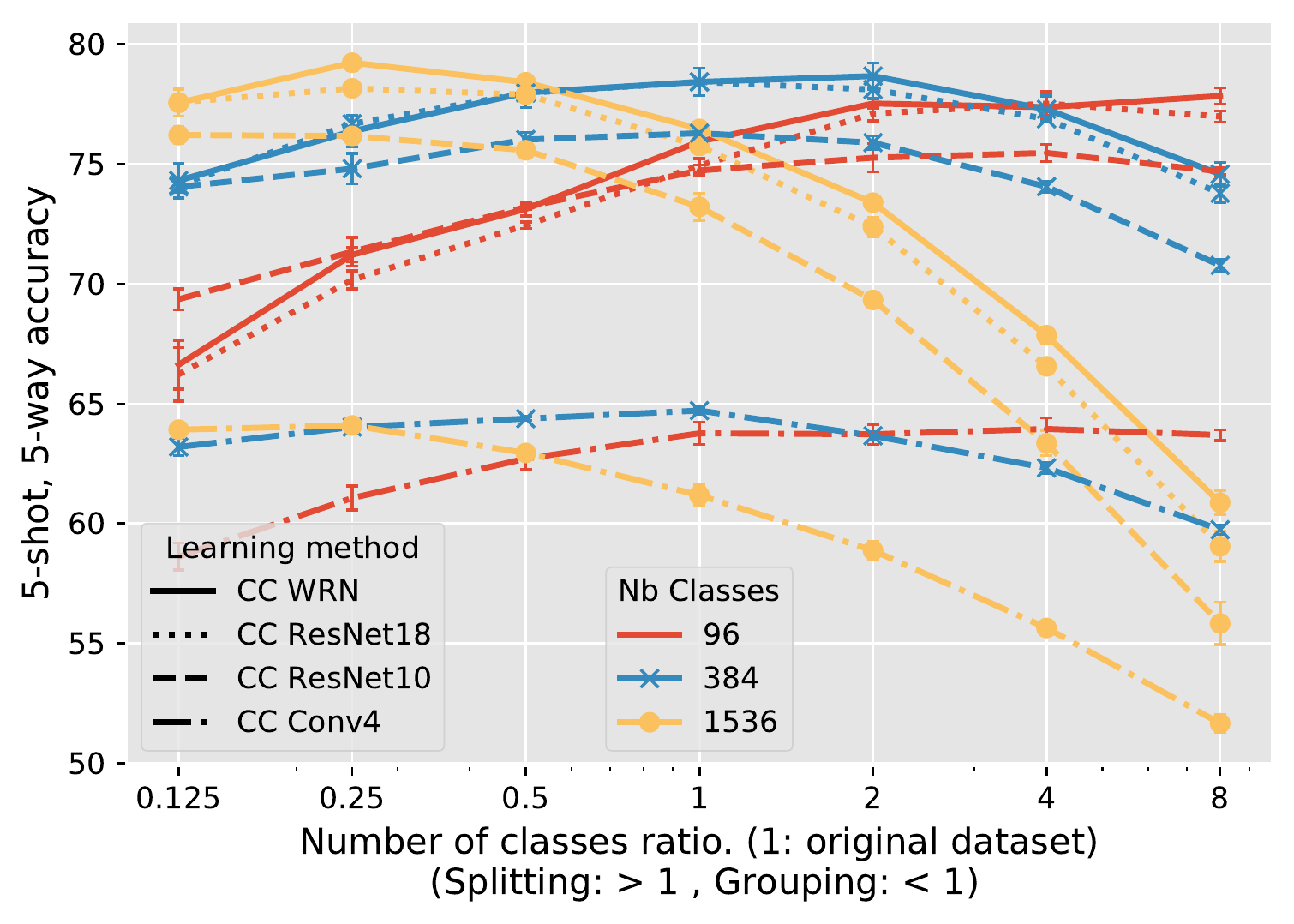} &
    \includegraphics[width=0.48\linewidth ]{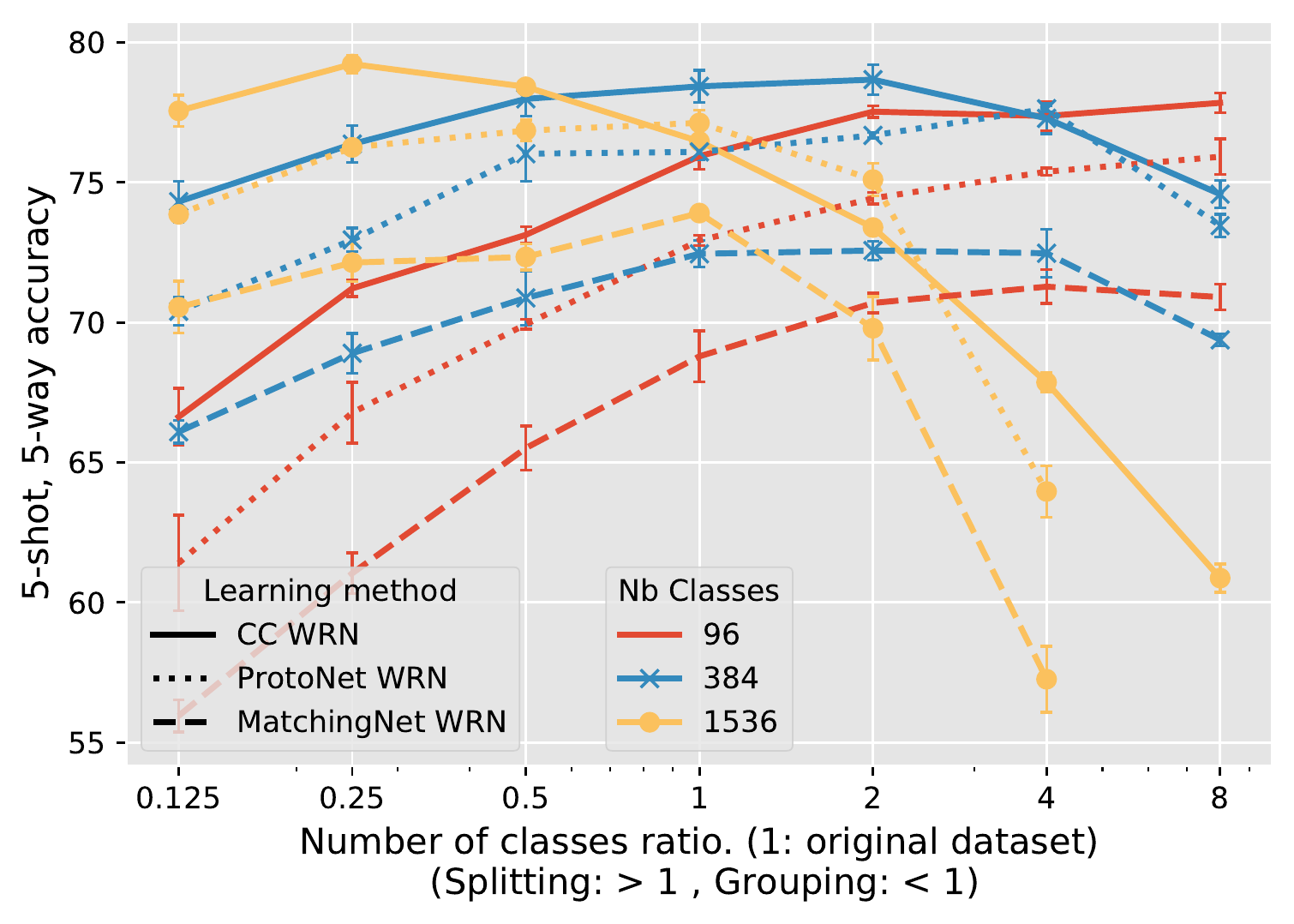} \\
    c) MiniIN, using different backbones & d) MiniIN, using different algorithms \\
    \end{tabular}
    \caption{Impact of class {grouping or splitting} on few-shot accuracy on miniIN and CUB depending on the initial number of classes. Starting from different number of classes $C$, we group similar classes together into meta-classes or split them into sub-classes to obtain $\alpha \times C$ ones. $\alpha \in \{\frac{1}{8}, \frac{1}{4}, \frac{1}{2}, 1, 2, 4, 8\}$ is the x-axis. Experiments in a) and b) use CC WRN setup.}
    \label{fig:splitting_grouping_classes}
\end{figure}

\subsection{Redefining classes}

There are two possible explanations for the improvement provided by the increased number of classes for a fixed number of annotated images discussed in the previous paragraph. The first one is that the images sampled from more random classes cover better the space of natural images, and thus provide images more likely similar to the test images. The second one is that learning a classifier with more classes is itself beneficial to the quality of the features. 
To investigate whether for fixed data, increasing the number of classes can boost performances, we \ccc{relabel} images inside each class as described in Section \ref{sec:splittinggrouping}.

In Figure~\ref{fig:splitting_grouping_classes}, we compare the effect of  grouping and splitting classes on three dataset configurations sampled from miniIN-6K and miniIN6K*, with a total number of images of 38400 with different number of classes $C \in \{96,384,1536\}$ for miniIN {and $C \in \{384,1536, 5704\}$ for CUB.}
Given images originally labeled with $C$ classes, we relabel images of each class to obtain $\alpha \times C$ sub-classes. 
The x-axes represent the class ratio $\alpha \in \{\frac{1}{8}, \frac{1}{4}, \frac{1}{2}, 1, 2, 4, 8\}$. For class ratios lower than $1$, we group classes using our greedy iterative grouping, while for ratios $\alpha$ greater than 1, we split classes using our BCP method.
In Fig~\ref{fig:splitting_grouping_classes} (a,b), we show three possible behaviors on miniIN and CUB when using our oracle features: (i) if the number of initial classes is higher than the optimal trade-off, grouping is beneficial and splitting hurts performances (yellow curves); (ii) if the number of initial classes is the optimal one, both splitting and grouping decrease performances (blue curves); (iii) if the number of initial classes is smaller than the optimal tradeoff, splitting is beneficial and grouping hurts performance (red curves). This is a strong result, since it shows there is potential to improve performances with a fixed training dataset by redefining new classes. This can be done for grouping using the self-supervised MoCo features. However, we found it is not sufficient to split classes in a way that improves performances. Using random features on the contrary does not lead to any significant improvements. 
Fig.~\ref{fig:splitting_grouping_classes}c confirms the consistency of results with various architecture on miniIN benchmark. Fig.~\ref{fig:splitting_grouping_classes}d compares these results to the ones obtained with ProtoNet and MatchingNet. Interestingly, we see that since the trade-off for these methods is with much fewer images per class, class splitting increases performances in all the scenarios we considered.

These results outline the need to adapt not only the base training images but also the base training granularity to the target few-shot task and algorithm. They also clearly demonstrate that the performance improvements we observe compared to standard trade-offs by using more classes and less images per class is not only due to the fact that the training data is more diverse, but also to the fact that training a classifier with more classes leads to improved features for few-shot classification.

\subsection{Selecting classes based on their diversity or difficulty}

\begin{figure}[htb]
    \centering
    \subfloat[\scriptsize{Class diversity on miniIN and CUB}]{
    \includegraphics[width=0.46\linewidth]{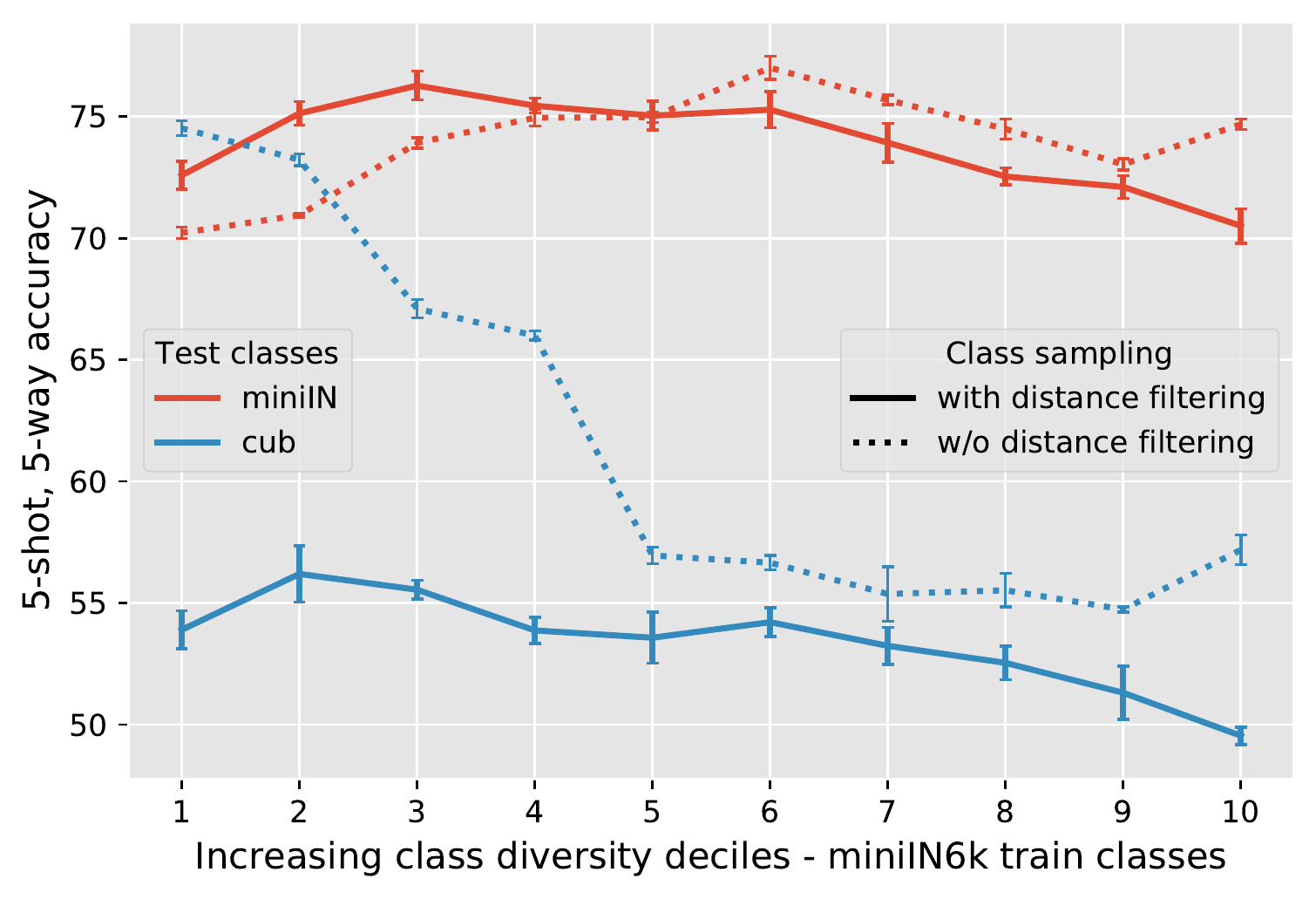}
    \label{fig:class_diversity_miniIN_cub}
    }
    \onlyarxiv{
    \subfloat[\scriptsize{Class diversity on miniIN1k diversity deciles}]{
    \includegraphics[width=0.48\linewidth]{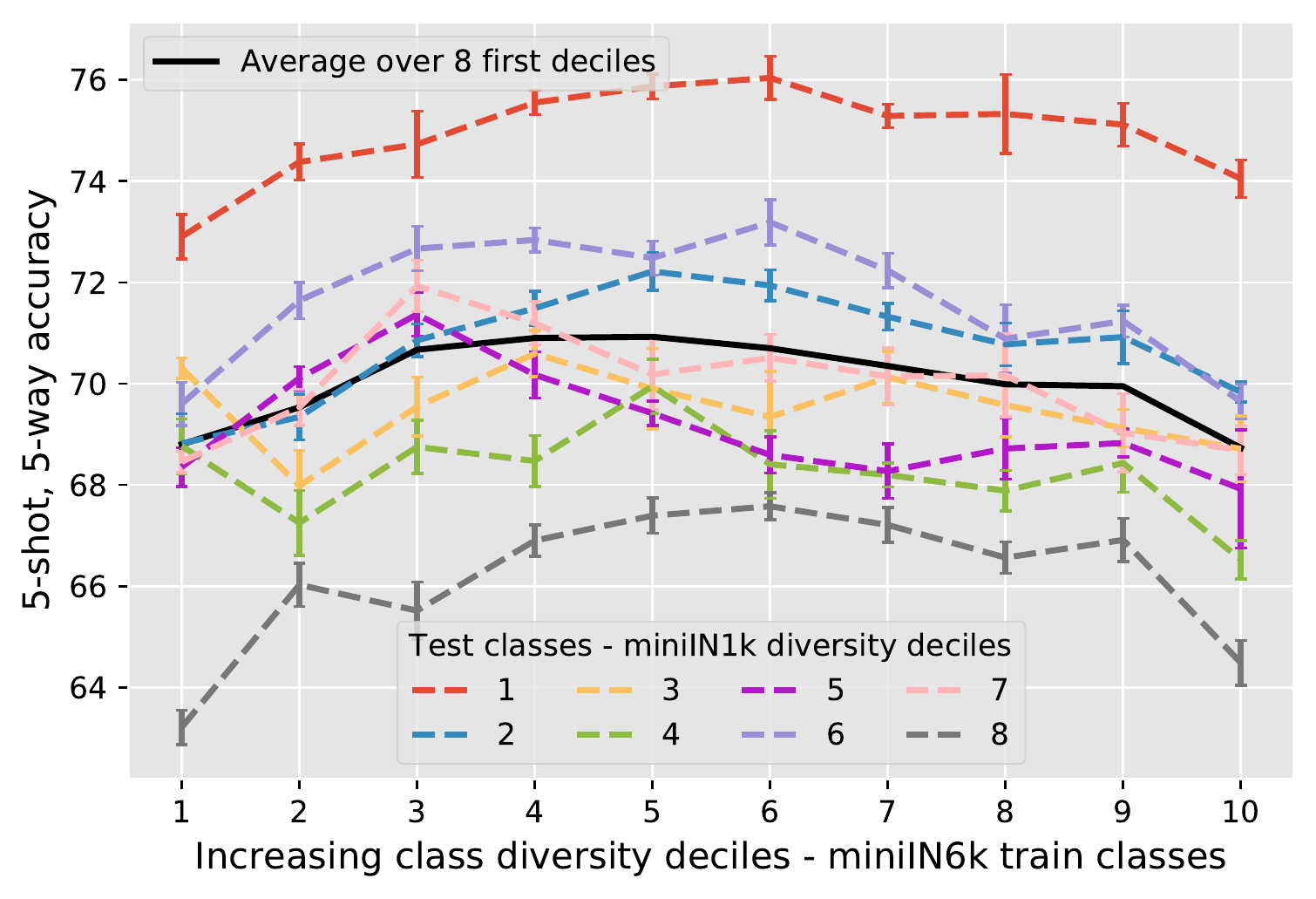}
    \label{fig:class_diversity_miniIN1k_deciles}}\\}
    \subfloat[\scriptsize{Val. acc. on miniIN and CUB}]{
    \includegraphics[width=0.46\linewidth]{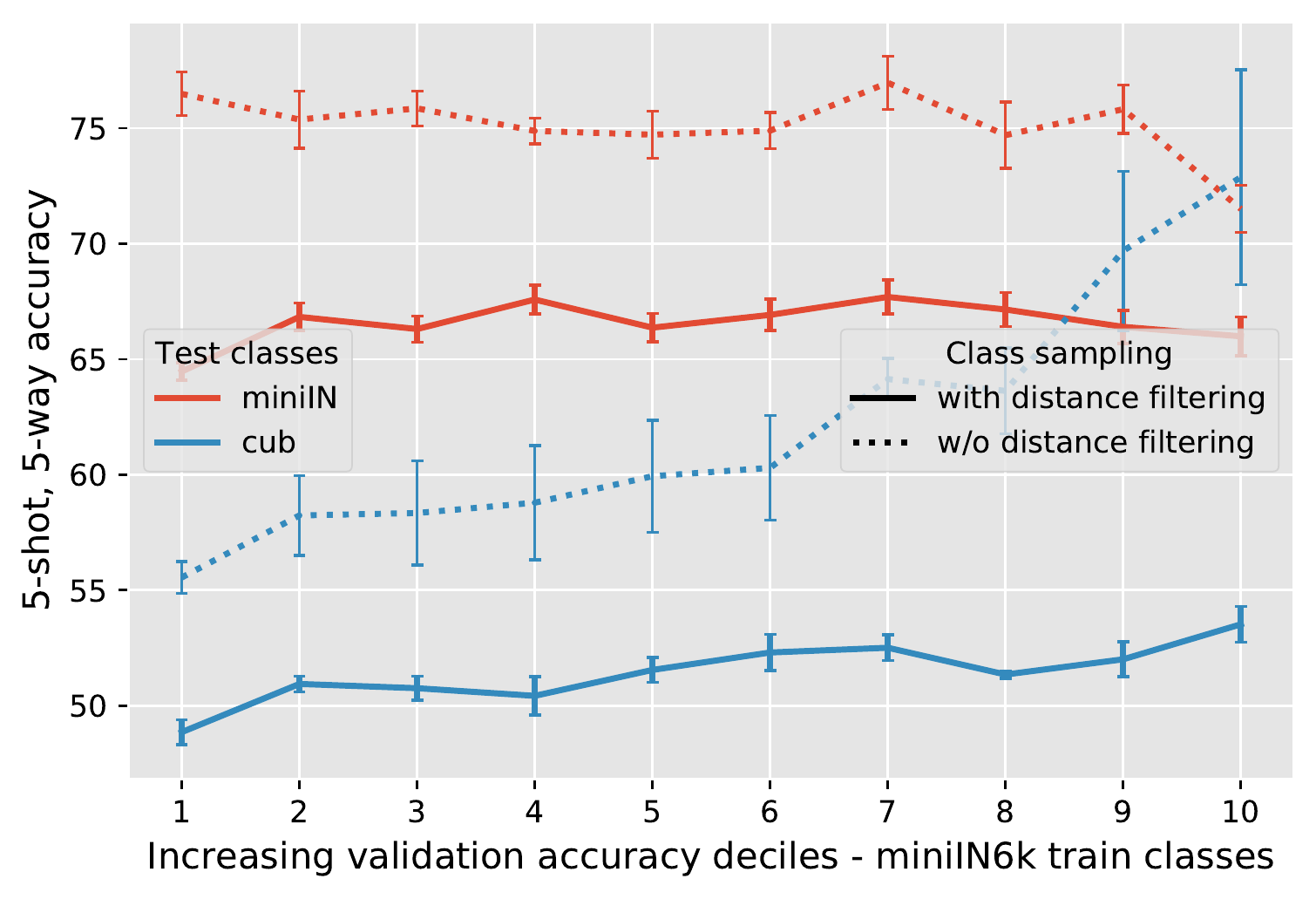}
    \label{fig:val_acc_miniIN_cub}
    }
    \onlyarxiv{\subfloat[\scriptsize{Val. acc. on miniIN1k val. acc. deciles}]{
    \includegraphics[width=0.48\linewidth]{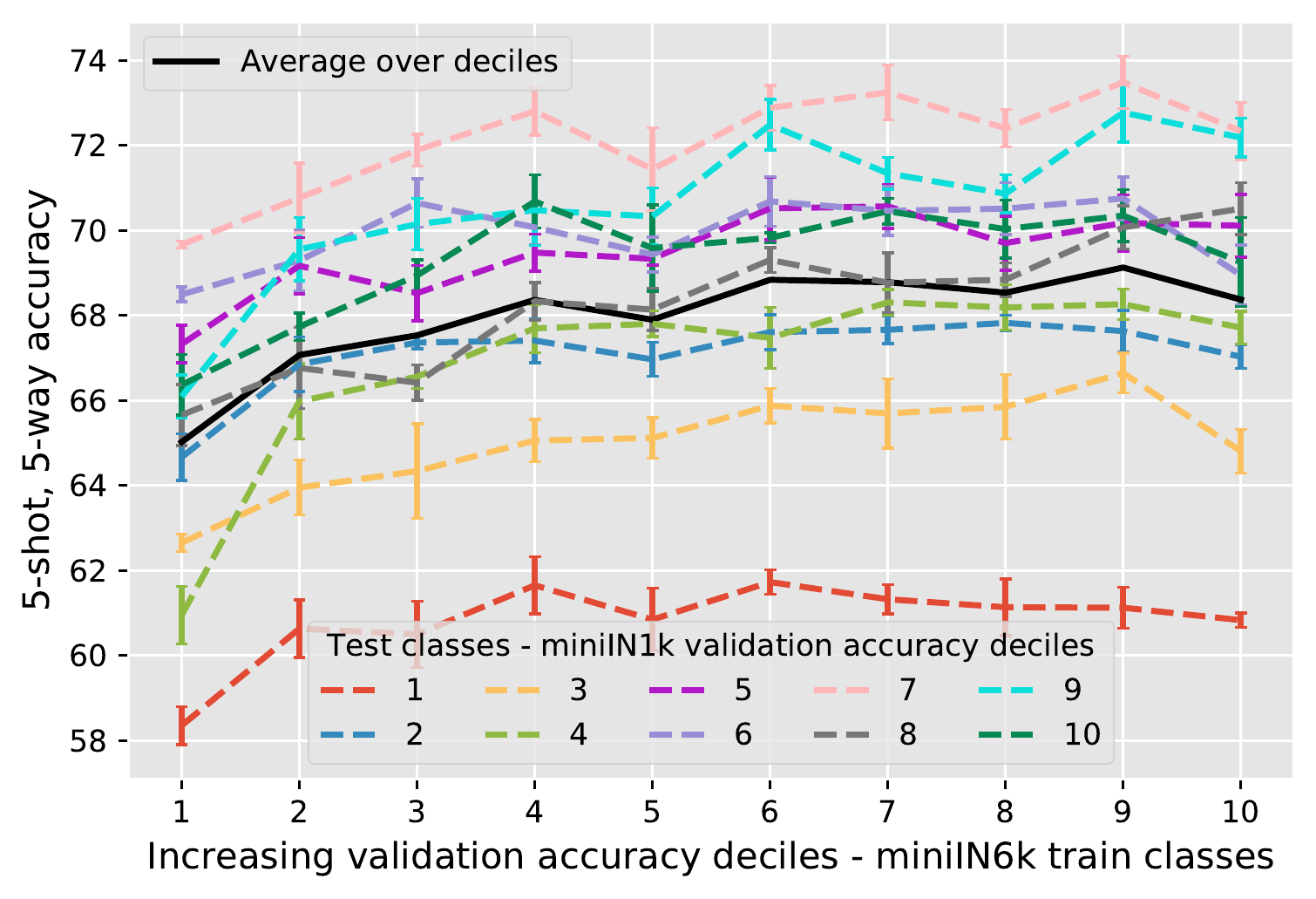}
    \label{fig:val_acc_miniIN1k_deciles}}\\}

    \caption{Impact of \textbf{class selection using class diversity and validation accuracy} on few-shot accuracy on miniIN and CUB\onlyarxiv{ and benchmarks sampled from miniIN-1K}. For training, we rank the classes of miniIN-6K in increasing feature variance or validation accuracy and split them into 10 bins from which we sample $C = 64$ classes that we use for base training. Fig. ~\ref{fig:class_diversity_miniIN_cub}, \ref{fig:val_acc_miniIN_cub} show the importance of selecting classes in each bin while considering their distance to test classes to disentangle both selection effects. \onlyarxiv{Fig. b) and d) show impact of class selection method on different benchmarks from miniIN1k sampled as deciles of increasing class diversity or validation accuracy.}}
    \label{fig:class_selection}
\end{figure}

After observing in Sec.~\ref{sec:sim} the importance of the similarity between base training classes and the test classes, we now study whether the diversity of the base classes or their difficulty is also an important factor. To this end, we compute the measures described in Sec.~\ref{sec:div} for every miniIN-6K class and rank them by increasing order. Then, we split the ranked classes into 10 bins of similar diversity or validation accuracy. The classes in the obtained bins are correlated to the test classes and thus introduces a bias in the performance due to this similarity instead of the diversity or difficulty we want to study (see \onlyarxiv{Figure~\ref{fig:class_diversity_similarity_corr} in }the \app, showing the similarity of classes in each bin to the test classes). To avoid this sampling bias, we associate to each class its distance to test classes, and sample base classes in each bin only in a small range of similarities, so that the average distance to the test classes is constant over all bins. In Fig.~\ref{fig:class_selection} we show the performances obtained by sampling using this strategy 64 classes and 600 images per class for a total of 38400 images in each bin. 
The performances obtained are shown on miniIN and CUB in Fig.~\ref{fig:class_diversity_miniIN_cub}, \ref{fig:val_acc_miniIN_cub} both using random sampling from the bin and using sampling with distance filtering as explained before. It can be seen that the effect of distance filtering is very strong, decreasing significantly the range of performance variation especially on the CUB dataset, however the difference in performance is still significant, around 5\% in all experiments. Both for CUB and miniIN, moderate class diversity - avoiding both the most and least diverse classes -  seems beneficial, while using the most difficult classes seems to harm performances. 
\onlyarxiv{To validate and provide additional insight on this experiment, we also use test benchmarks sampled from miniIN1k with classes grouped by their diversity or validation accuracy deciles from 1 to 10 in Fig.~\ref{fig:class_selection} (b, d). The curve in black shows the average over all the bins. While the range of performances highly depends on the test class selection criteria, the tendency seem very consistent on each of them. For class diversity, we observe a inverted U shape average curve, {i.e.} using most or least diverse classes can hurt the few-shot performance, with optimal performances corresponding to slightly lower than average diversity. For validation accuracy, better few-shot classification performance is correlated with higher class validation performance, i.e. using classes that are easier to classify lead to better feature for few-shot classification.}

\section{Conclusion}
Our empirical study outlines the key importance of the base training data in few-shot learning scenarios, with seemingly minor modifications of the base data resulting in large changes in performance, and carefully selected data leading to much better accuracy. We also show that few-shot performance can be improved by automatically relabelling an initial dataset by merging or splitting classes. We hope the analysis and insights that we present will:
\begin{enumerate}[wide, labelwidth=!, labelindent=0pt,noitemsep,topsep=0pt]
\item impact dataset design for practical applications, e.g. given a fixed number of images to label, one should prioritize a large number of different classes and potentially use class grouping strategies using self-supervised features. In addition to base classes similar to test data, one should also prioritize simple classes, with moderate diversity. 
\item lead to new evaluations of few-shot learning algorithm, considering explicitly the influence of the base data training in the results: the current miniIN setting of 64 classes and 600 images per class is far from optimal for several approaches. Furthermore, the optimal trade-off between number of classes and number of images per class is different for different few-shot algorithms, suggesting taking into account different base data distributions in future few-shot evaluation benchmarks.
\item inspire advances in few-shot learning, e.g. the design of practical approaches to adapt base training data automatically and efficiently to target few-shot tasks.
\end{enumerate}

\paragraph{Acknowledgements:}
This work was supported in part by ANR project EnHerit ANR-17-CE23-0008, project Rapid Tabasco.  We thank Maxime Oquab, Diane Bouchacourt and Alexei Efros for helpful discussions and feedback.

\bibliographystyle{splncs04}
\bibliography{main}

\onlyarxiv{
\clearpage
\appendix
\section{Appendix}
We provide in \onlyarxiv{this appendix} complementary 1-shot results of Table~\ref{tab:table1}, and results for class splitting on the ImageNet benchmark. We also present a comparison of the chosen relabeling algorithms to K-means and finally give details about the MiniIN-6k dataset and the used MoCo features.

\subsection{Results of using MiniIN-6k on different benchmark}
We report in Table~\ref{tab:table1_1shot}, complementary 1-shot results to the ones in table~\ref{tab:table1} on both miniImagenet and CUB datasets, using four different feature backbones. These results are consistent with the observations made on the impact of the base training data on the five shot accuracy. 

\begin{table}[htbp]
\begin{center}
\resizebox{\textwidth}{!}{
\begin{tabular}[htp]{cc|c|c|c|c|c|c|}
\cline {3-8}
& &  \multicolumn{3}{c|}{MiniIN } &\multicolumn{3}{c|}{CUB }\\
\cline {2-8} 
& \multicolumn{1}{|c|}{\backslashbox{Algo.}{Base\\data}}& \begin{tabular}{@{}c@{}} MiniIN \\ $N$=38400 \\ $C=64$ \end{tabular} & \begin{tabular}{@{}c@{}} MiniIN6K\\ Random\\$N$=38400  \\$C=64$ \end{tabular} & \begin{tabular}{@{}c@{}} MiniIN6K \\$N$$\approx$7,1.$10^6$\\ $C$=6000 \end{tabular} & \begin{tabular}{@{}c@{}} CUB \\  $N$=5885  \\ $C=100$ \end{tabular} & \begin{tabular}{@{}c@{}} MiniIN6K*\\ Random \\ $N$=38400  \\ $C=64$ \end{tabular} & \begin{tabular}{@{}c@{}} MiniIN6K* \\$N$$\approx$6,8.$10^6$\\ $C=5704$ \end{tabular} \\
\hline 
\multicolumn{1}{|l|}{WRN}& CC &
 \mvar{61.62}{0.17}  & \mvar{58.49}{2.29} &  \mvar{85.40}{0.15} & \mvar{76.73}{0.40}  & \mvar{41.62}{0.93} &  \mvar{73.51}{0.21} \\
 \hline
\multicolumn{1}{|l|}{Conv4}& CC & 
 \mvar{48.62}{0.09}  & \mvar{46.87}{0.70} &  \mvar{56.09}{0.16} & \mvar{61.21}{0.16}  & \mvar{39.65}{0.71} &  \mvar{47.01}{0.26} \\
\hline
\multicolumn{1}{|l|}{ResNet10}& CC &
 \mvar{59.06}{0.35}  & \mvar{56.06}{1.74} & \mvar{74.42}{0.20} & \mvar{74.48}{0.42} & \mvar{40.92}{0.51}  & \mvar{57.81}{0.43} \\
\hline
\multicolumn{1}{|l|}{ResNet18}& CC &
 \mvar{60.85}{0.17}  & \mvar{57.51}{1.79} &  \mvar{81.42}{0.20} & \mvar{76.13}{0.39}  & \mvar{40.90}{0.86} &  \mvar{63.14}{0.93} \\
\hline
\end{tabular}
} 
\end{center}
\caption{One shot, 5-way accuracy on MiniIN and CUB using different base training data and backbones. CC: Cosine Classifier. WRN: Wide ResNet28-10. MiniIN6K Random: 600 images from 64 classes sampled randomly from MiniIN6K. We evaluate the variances over 3 different runs, each run compute the few-shot performance on 10k sampled episodes. \ccc{MiniIN6K*: MiniIN6K without images from bird categories.}}

\label{tab:table1_1shot}
\end{table}

\subsection{Class selection with similarity to test classes}

Similarly to the Fig~\ref{fig:adding_classes}a, we show results of selecting closest or farthest classes to CUB test classes using different features in Fig~\ref{fig:cub_similarity_features}. Similar observations can be drawn.

\begin{figure}
    \centering
    \includegraphics[width=0.48\linewidth]{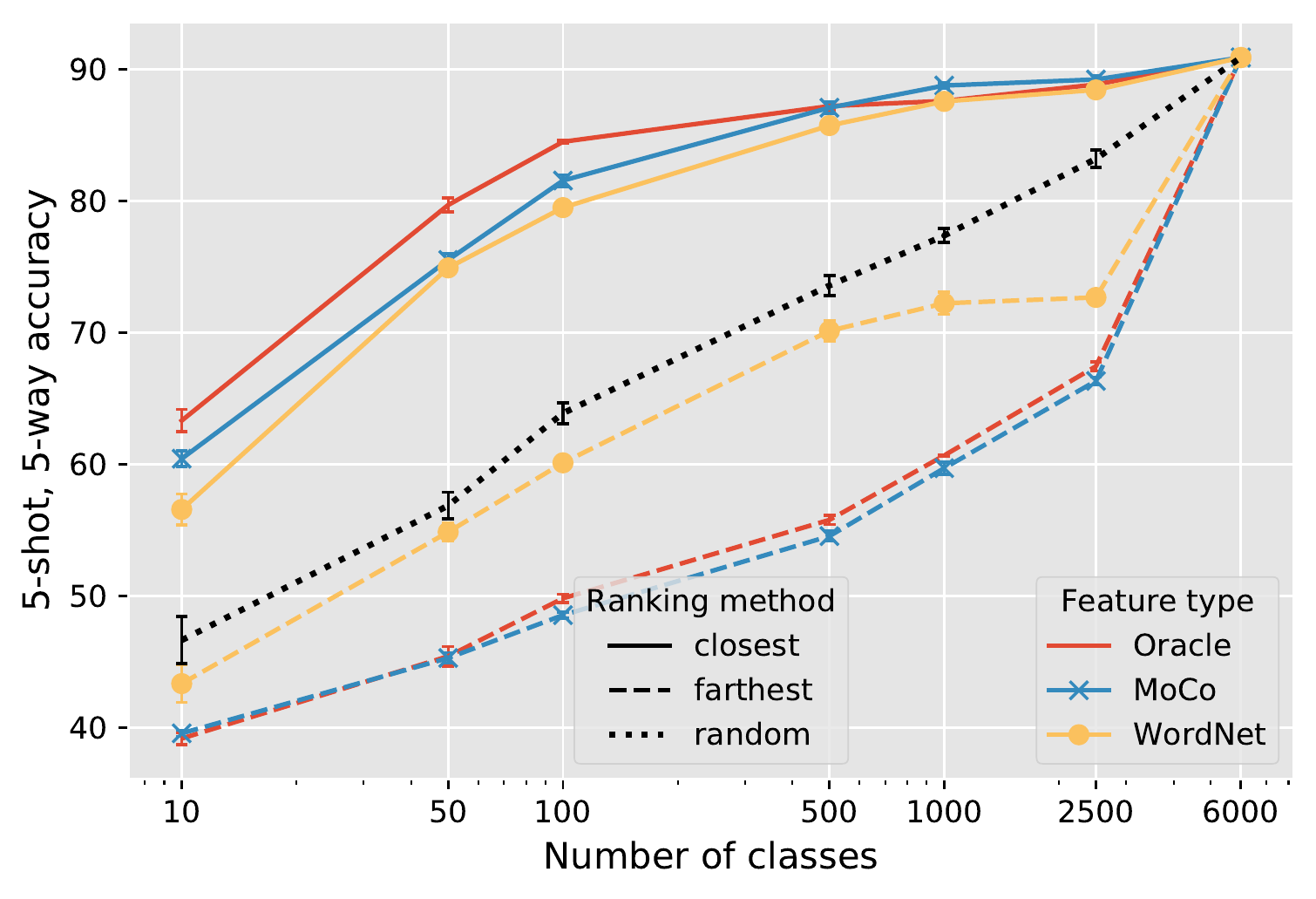}

    \caption{Five-shot accuracy on CUB when sampling classes from miniIN-6K {\bf closest/farthest} to the CUB test set or randomly. }
    \label{fig:cub_similarity_features}
\end{figure}

\subsection{Number of classes and images trade-off on IN benchmark}

Similarly to Fig~\ref{fig:tradeoff}, we show results of sampling datasets from IN6k of 500k or 50k images and with different number of classes.
Using 50k images, we observe that there is an optimal trade-off between the number of classes and number of images, while for 500k, the optimal number of classes might be larger than the maximum possible using IN6k (i.e 6000).

\begin{figure}
    \centering
    \includegraphics[width=0.48\linewidth]{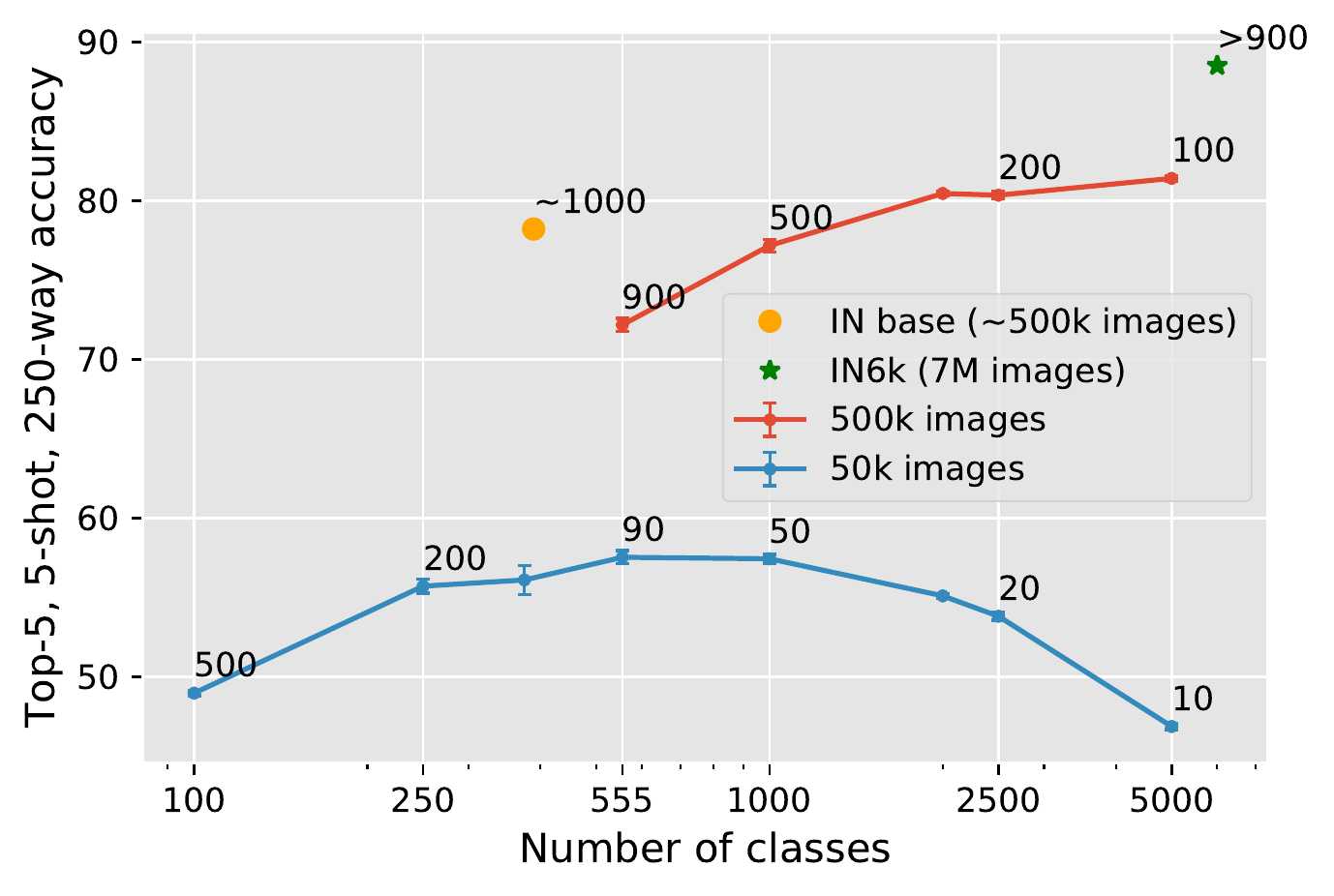}
    \caption{Trade-off between the number of classes and images per class for a fixed image budget on the IN benchmark. Each point is annotated with its corresponding number of images per class.}
    \label{fig:tradeoff_IN}
\end{figure}

In Table~\ref{tab:IN_bench}, we present results on the impact of splitting classes of the training dataset using oracle features on the ImageNet-1k high resolution benchmark as defined in~\cite{hariharan2017low}. Splitting classes improves the 5-shot performance by 2\% and the 1-shot performance by barely 0.93\% using ResNet-34.

\begin{table}
    \centering
    
    \begin{tabular}{c|c|c|c|c|c|}
    \cline{2-6}
                 & base & split 2 & split 4 & split 8 & IN6k \\ \midrule
    \multicolumn{1}{|c|}{1-shot}  & \mvar{57.56}{0.20} & \mvar{58.12}{0.00} & \mvar{58.49}{0.17} & \mvar{57.78}{0.01} & \mvar{70.94}{0.07} \\ \hline
    \multicolumn{1}{|c|}{5-shot}  & \mvar{78.15}{0.23} & \mvar{79.13}{0.03} & \mvar{80.10}{0.15} & \mvar{80.30}{0.07} & \mvar{88.50}{0.05} \\ \bottomrule
    \end{tabular}
    
    \caption{Top-5, 250-way few shot accuracy on the ImageNet-1k benchmark using the 389 base training classes - 497350 images (base), and split versions of this dataset into 2,4 and 8 splits and also using the large IN6k dataset (7135118 images).  Results are averaged over 3 different runs.}
    \label{tab:IN_bench}
\end{table}

\subsection{Comparing splitting and grouping strategies to K-means}

In Figure \ref{fig:cluster_methods}, we show how the used splitting and grouping methods compare to a simple K-means algorithm (leading to unbalanced clusters). We observe that the balanced class relabeling leads to a better performance than the K-means based relabeling, justifying our choice for the presented experiments. These results use oracle features to compute image features for class splitting.

\begin{figure}
    \centering
    \begin{tabular}{cc}
    \includegraphics[width=0.49\linewidth]{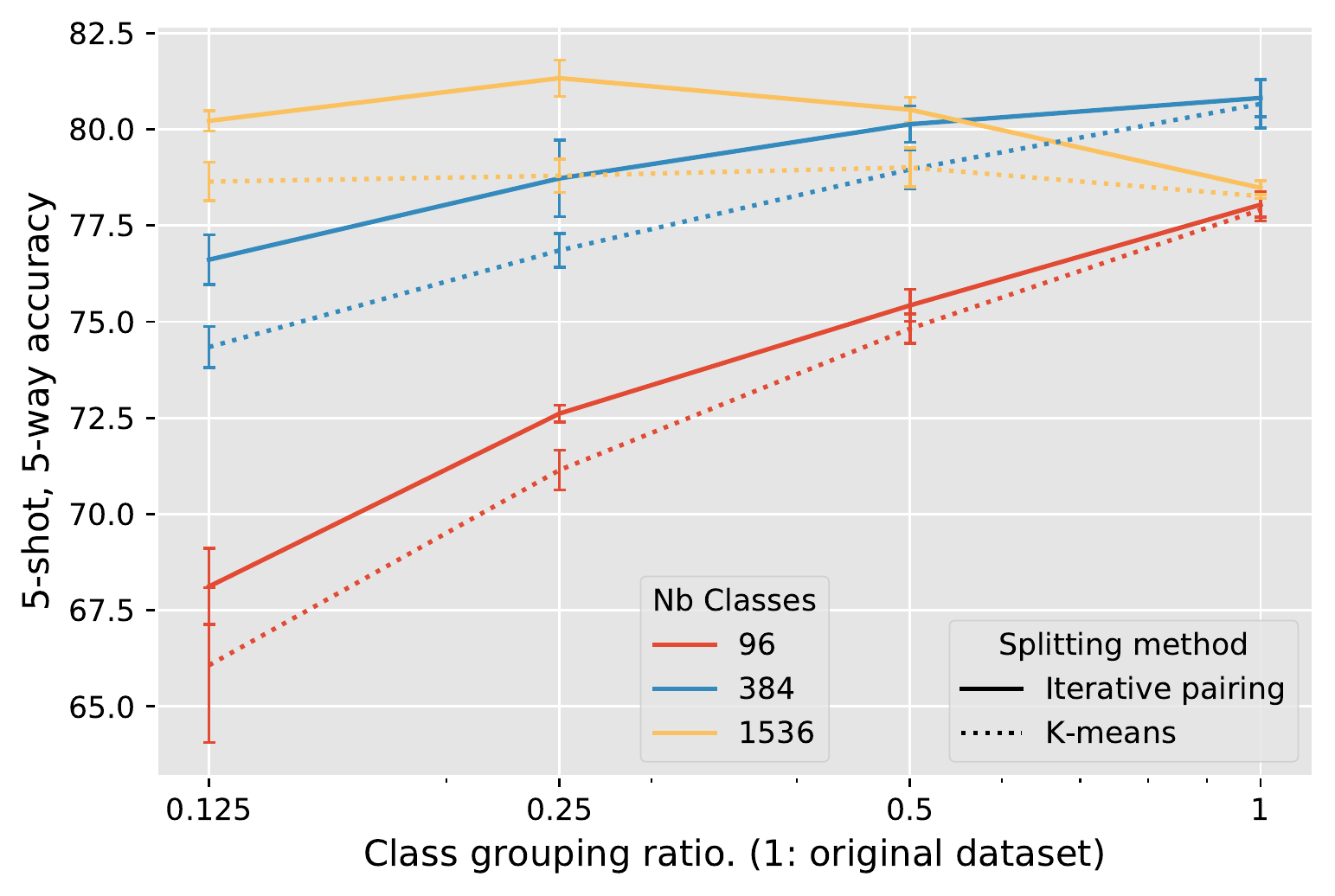} &
    \includegraphics[width=0.49\linewidth]{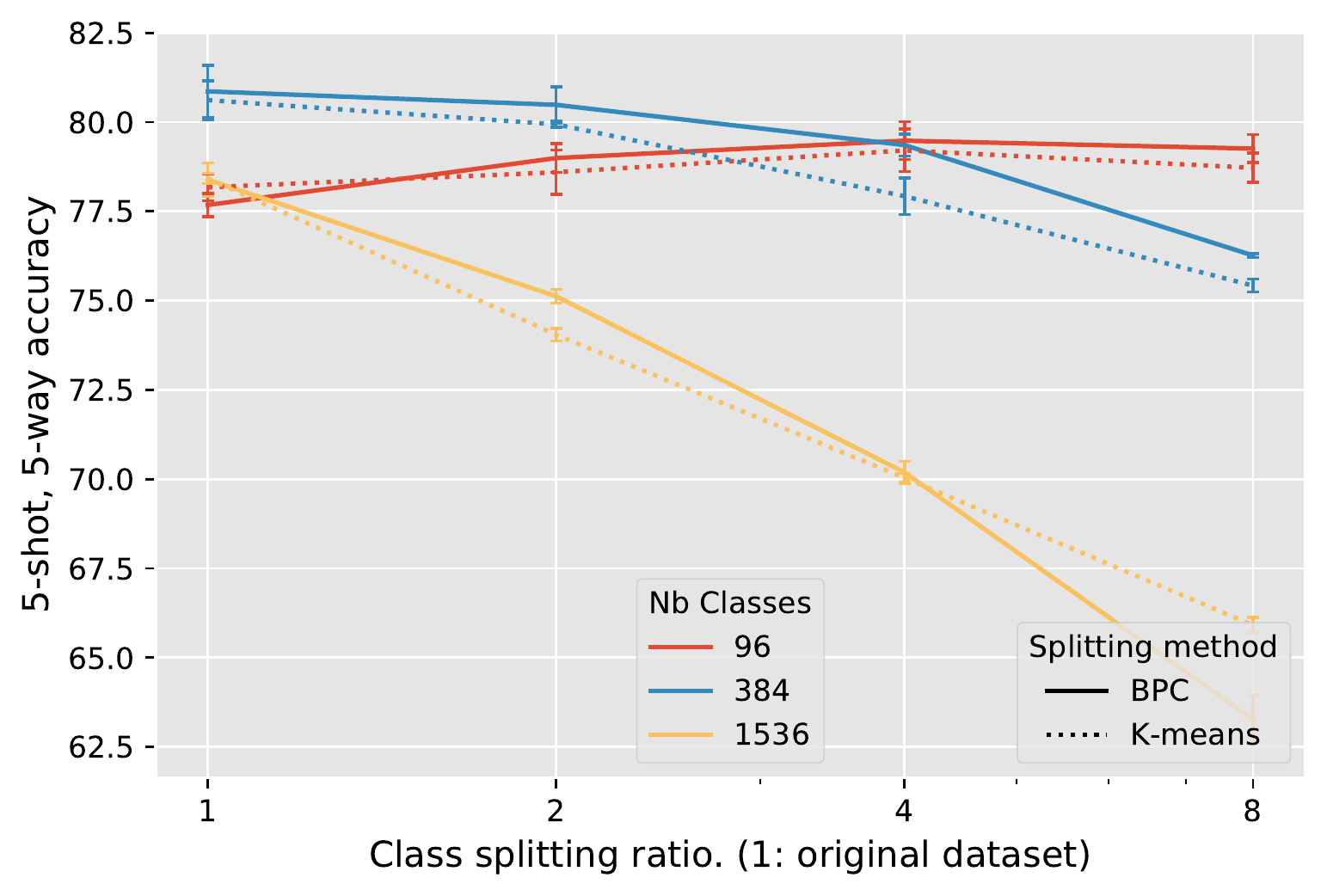} 
    \end{tabular}
    \caption{Comparing our balanced splitting and grouping methods (Bisection along Principal Component for splitting classes, and Iterative pairing for grouping classes) to K-means on the MiniIN benchmark.}
    \label{fig:cluster_methods}
\end{figure}

\subsection{Details about MoCo features}
We use the self-supervised features on ImageNet using a ResNet-50 backbone from \cite{tian2019contrastive}\footnote{Moco features from \url{https://github.com/HobbitLong/CMC/}} unofficial implementation of Momentum Contrast for unsupervised visual representation learning~\cite{he2019momentum}.

\subsection{Details about miniIN6K}
We created the ImageNet-6K dataset by sampling the largest 6000 classes from ImageNet-22K excluding the Imagenet-1K classes. Each class contains at least 900 images and a maximum of 2248 available images per class with a total of 7135116 images. We will share the list of images in the IN6K dataset.

\subsection{Class sampling bias}
\os{In Fig.~\ref{fig:class_diversity_similarity_corr}, we show the correlation between the distance to miniIN test classes of miniIN6k classes grouped into 10 bins of increasing class diversity or validation performance. We observe that most diverse classes are closest to miniIN test classes than least diverse ones.}

\begin{figure}
    \centering
\begin{tabular}{cc}
    \includegraphics[width=0.49\linewidth]{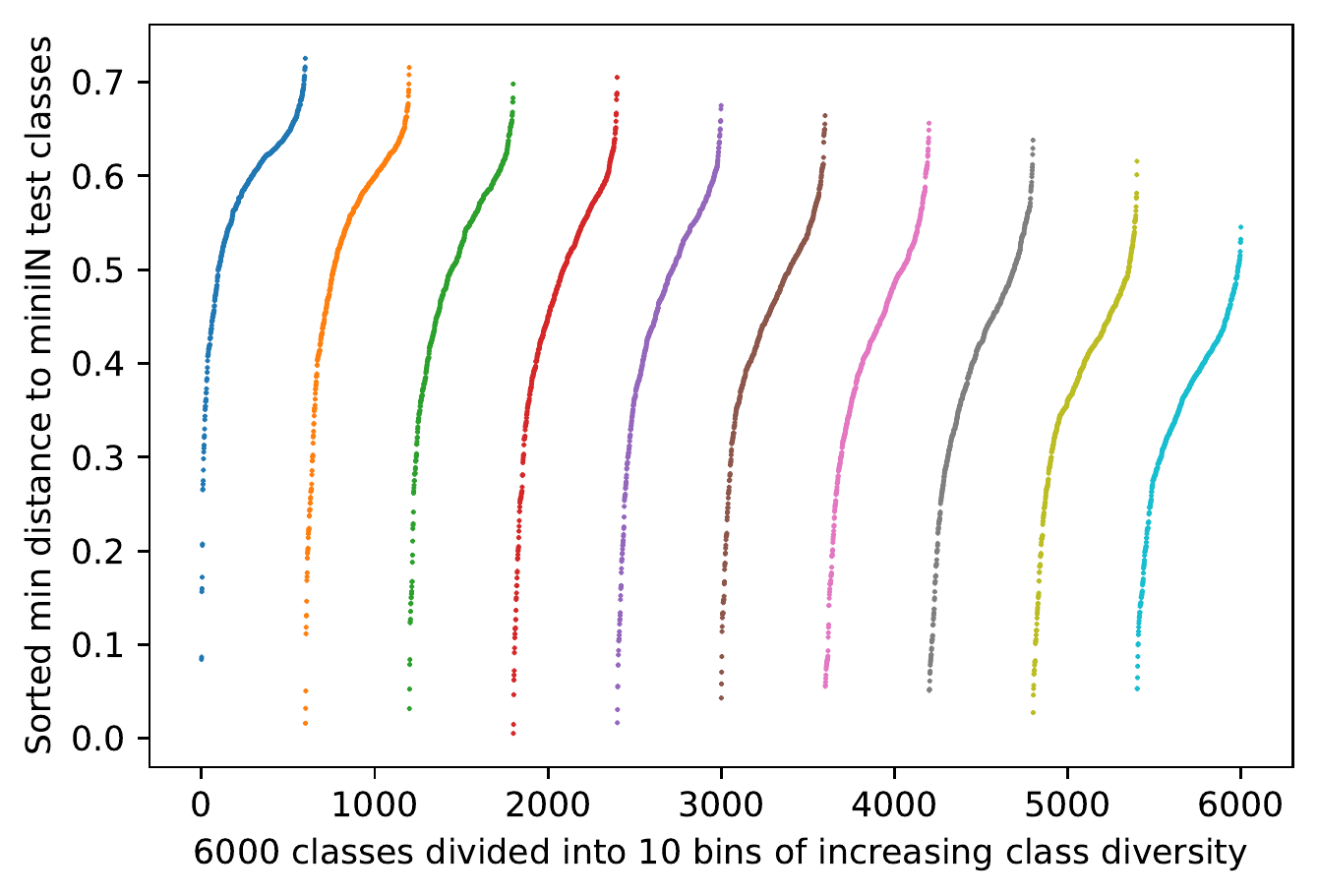} &  
    \includegraphics[width=0.49\linewidth]{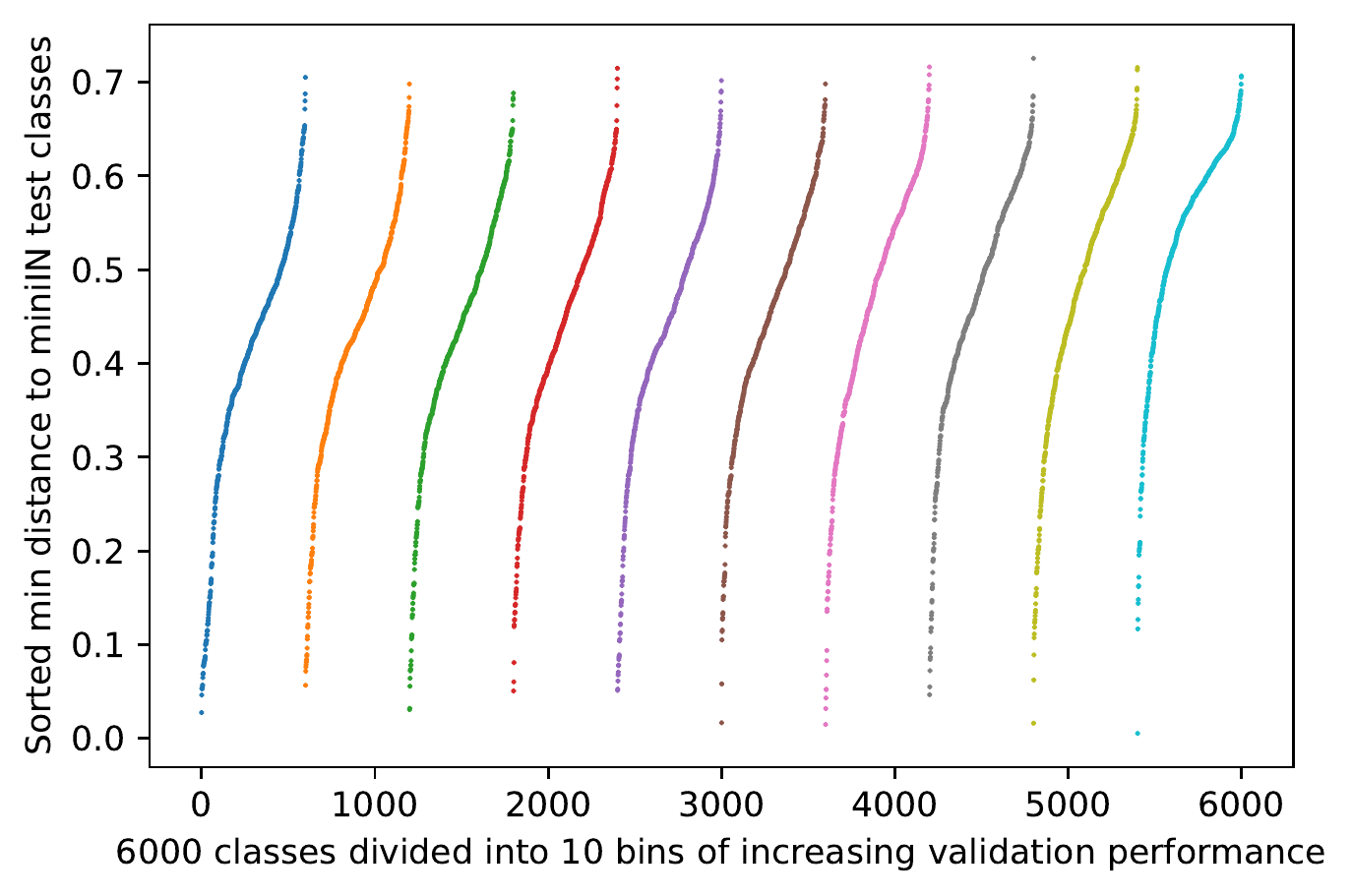}
\end{tabular}

    \caption{Class similarity between miniIN6k classes and miniIN test classes. MiniIN6k classes  (x-axis) are grouped in 10 bins of increasing class diversity or validation accuracy. We observe that class similarity to test classes correlates with both class diversity and class validation accuracy, thus the importance of avoiding this bias during class selection.}
    \label{fig:class_diversity_similarity_corr}
\end{figure}

\subsection{Image credits of Figure~\ref{fig:class_splitting}}
Sunflower and tulip images are licensed under the Creative Commons By-Attribution License, available at:
\url{https://creativecommons.org/licenses/by/2.0/}
The photographers are listed below, thanks to all of them for making their work available.

\resizebox{\textwidth}{!}{
\begin{tabular}{l}
    1240625276\_fb3bd0c7b1.jpg CC-BY by Rob Young - https://www.flickr.com/photos/rob-young/1240625276/\\
    2443921986\_d4582c123a.jpg CC-BY by Ally Aubry -     https://www.flickr.com/photos/allyaubryphotography/2443921986/\\
    58636535\_bc53ef0a21\_m.jpg CC-BY by sophie \& cie -     https://www.flickr.com/photos/biscotte/58636535/\\
    3062794421\_295f8c2c4e.jpg CC-BY by liz west -     https://www.flickr.com/photos/calliope/3062794421/\\
    5994572653\_ea98afa3af\_n.jpg CC-BY by Manu -     https://www.flickr.com/photos/seven\_of9/5994572653/\\
    8174972548\_0051c2d431.jpg CC-BY by Stephane Mignon -     https://www.flickr.com/photos/topsteph53/8174972548/\\
    3568925290\_faf7aec3a0.jpg CC-BY by cbransto -     https://www.flickr.com/photos/cbransto/3568925290/\\
    14460081668\_eda8795693\_m.jpg CC-BY by Forsaken Fotos -     https://www.flickr.com/photos/55229469@N07/14460081668/\\
    405035580\_94b793e71d.jpg CC-BY by ethan lindsey - https://www.flickr.com/photos/ethanlindsey/405035580/\\
    3990989735\_59e2751151\_n.jpg CC-BY by Allie\_Caulfield - https://www.flickr.com/photos/wm\_archiv/3990989735/\\
    4838669164\_ffb6f67139.jpg CC-BY by Erik Eskedal - https://www.flickr.com/photos/eskedal/4838669164/\\
    8710148289\_6fc196a0f8\_n.jpg CC-BY by liz west - https://www.flickr.com/photos/calliope/8710148289/\\
    11746276\_de3dec8201.jpg CC-BY by Dan Kamminga - https://www.flickr.com/photos/dankamminga/11746276/\\
    2440874162\_27a7030402\_n.jpg CC-BY by Jon - https://www.flickr.com/photos/jonparry/2440874162/\\
    5012813078\_99fb977616\_n.jpg CC-BY by Roel Hemkes - https://www.flickr.com/photos/rhemkes/5012813078/\\
    13510068773\_c925c5517c.jpg CC-BY by nikontino - https://www.flickr.com/photos/nikontino/13510068773/\\
\end{tabular}
}
}

\end{document}